\documentclass[letterpaper, 10 pt, conference]{ieeeconf}  

\IEEEoverridecommandlockouts                              

\usepackage{graphics} 
\usepackage{amsmath} 
\usepackage{amssymb}  
\usepackage{graphicx}
\usepackage{pifont}
\usepackage[table]{xcolor}
\usepackage{multirow}
\usepackage{threeparttable}

\title{\LARGE \bf
EndoUFM: Utilizing Foundation Models for Monocular depth estimation of endoscopic images
}

\author{Xinning Yao$^{1}$, Bo Liu$^{1, 2}$, Bojian Li$^{1}$, Jingjing Wang$^{1}$, Jinghua Yue$^{1}$, and Fugen Zhou$^{1}$ 
\thanks{This work was supported in part by the Beijing Natural Science Foundation (L232037, L242112, and L252005).}
\thanks{$^{1}$Image Processing Center, Beihang University, Beijing, 100191, China. $^{2}$State Key Laboratory of High-Efficiency Reusable Aerospace Transportation Technology, Beijing 102206, China. Corresponding Author: {\tt\small bo.liu@buaa.edu.cn}}%
}

\begin{document}

\maketitle
\thispagestyle{empty}
\pagestyle{empty}

\begin{abstract}
Depth estimation is a foundational component for 3D reconstruction in minimally invasive endoscopic surgeries. However, existing monocular depth estimation techniques often exhibit limited performance to the varying illumination and complex textures of the surgical environment. While applying foundation models offers a promising approach to enhance the depth estimation performance, the domain gap between the natural images used for pre-training and the target endoscopic images leads to significant semantic perception deficiencies. In this study, EndoUFM is introduced as an unsupervised monocular depth estimation framework that innovatively \underline{U}tilizes dual \underline{F}oundation \underline{M}odels for \underline{Endo}scopic images, thereby enhancing the depth estimation performance by leveraging the powerful pre-learned priors. The framework features a novel adaptive fine-tuning strategy that incorporates Random Vector Low-Rank Adaptation (RVLoRA) to enhance model adaptability, and a Residual block based on Depthwise Separable Convolution (Res-DSC) to improve the capture of fine-grained local features. A mask-guided smoothness loss is also introduced to enforce depth consistency within anatomical structures. Extensive experiments on the SCARED, Hamlyn, SERV-CT, and EndoNeRF datasets confirm that our method achieves state-of-the-art performance while maintaining an efficient model size. This work contributes to augmenting surgeons' spatial perception during minimally invasive procedures, thereby enhancing surgical precision and safety, with crucial implications for augmented reality and navigation systems. Our code is available at https://github.com/RealMindyY/EndoUFM.
\end{abstract}




\maketitle


\section{Introduction}

Minimally Invasive Surgery (MIS) has revolutionized modern medicine, offering patients smaller incisions and quicker recovery times. To further advance surgical precision and safety, the field is increasingly turning to robotic systems enhanced with augmented reality (AR) \cite{sharma2025utilization, yang2024selflightweight}. A cornerstone of these AR applications is the ability to generate a real-time 3D reconstruction of the surgical site, which fundamentally relies on accurate depth estimation. Given the challenges in obtaining ground-truth data from inside the body and the widespread use of single-camera endoscopes, unsupervised monocular depth estimation (UMDE) has emerged as a particularly flexible and practical area of research \cite{liu2024self, zhou2017unsupervised}.

Despite this flexibility, the endoscopic environment presents a formidable challenge for existing depth estimation techniques. The confined surgical cavity is subject to dynamic and complex conditions, including non-uniform illumination, intense specular reflections off tissue surfaces, tissue deformation, and frequent occlusions by surgical tools. These factors severely undermine traditional UMDE methods, which often rely on photometric consistency and rigid scene assumptions that do not hold up in a live surgical setting \cite{ozyoruk2021endoslam, li2024image, godard2019digging}. While some strategies have been developed to address these issues, such as modeling illumination changes \cite{shao2022self} or using intrinsic image decomposition \cite{li2024image}, they still struggle with the drastic lighting shifts and limited global perception inherent in traditional Convolutional Neural Network (CNN) architectures.

Meanwhile, the recent advent of large-scale foundation models offers a powerful new path forward \cite{han2024depth}. Models like Depth Anything Model \cite{yang2024depth}, pre-trained on millions of diverse natural images, have learned robust geometric priors that are invaluable for understanding 3D space. Similarly, medical-specific models like MedSAM \cite{ma2024segment}, an adaptation of the Segment Anything Model (SAM) , excel at identifying and outlining complex anatomical structures in medical imagery. Harnessing these powerful, pre-learned priors can dramatically improve depth estimation in challenging endoscopic images.

However, despite this tremendous potential, how to fully and efficiently utilize the pre-trained priors of different foundation models still need to be explored. Current methods typically utilize only one particular foundation model \cite{cui2024surgical,cui2024endodac,yang2024selfCLIP}, failing to systematically combine the geometric and semantic priors for enhancing UMDE. Furthermore, the domain gap between natural pre-training images and target endoscopic images causes significant semantic perception deficiencies \cite{han2024depth}. Though there are parameter-efficient fine-tuning methods (such as the noted Low-Rank Adaptation (LoRA) \cite{hu2022lora} ) to handle this, they often lack the robustness to overcome the following fundamental endoscopic challenges. First, they struggle with the inherent scale ambiguity of monocular vision \cite{Bian2019unsupervisedscale,Song2024unsupervisedconsistency,ZHANG2026Multi}. This ambiguity makes it difficult to perceive the actual size and distance of objects from a single 2D image, as a small, nearby object can appear identical to a large, distant one. Second, standard adaptation methods frequently fail to adequately capture the local, fine-grained details characteristic of complex endoscopic scenes.

To address this gap, this work proposes EndoUFM as a novel unsupervised monocular depth estimation framework of \underline{Endo}scopic images that synergistically \underline{U}tilizes dual \underline{F}oundation \underline{M}odels. The decomposition-based UMDE framework \cite{li2024image} consists of two core components: a depth estimation network and an image decomposition module, which correspond to the geometric representation of the Depth Anything Model and the semantic representation of the Segment Anything Model (SAM), respectively. Inspired by this alignment, EndoUFM is designed to jointly exploit these powerful pre-trained priors and improve depth estimation under challenging endoscopic conditions. To overcome domain adaptation difficulties, a sophisticated adaptive fine-tuning strategy is introduced. This strategy features Random Vector Low-Rank Adaptation (RVLoRA), which introduces random perturbations to encourage scale-robust representation learning and mitigate scale ambiguity. Additionally, Residual blocks based on Depthwise Separable Convolution (Res-DSC) are incorporated to capture fine-grained local features essential for precise depth prediction. To further ensure semantic consistency, a mask-guided smoothness loss is designed to enforce depth consistency within anatomical tissue structures using SAM-derived segmentation masks. 

Extensive experiments on the SCARED, Hamlyn, SERV-CT, and EndoNeRF datasets demonstrate that EndoUFM achieves state-of-the-art performance, surpassing existing methods in robustness and accuracy. By enhancing surgeons’ spatial perception, EndoUFM holds transformative potential for improving AR and navigation systems in MIS, thereby advancing surgical precision and safety. 

In summary, the main contribution of this work can be summarized as the following: 
\begin{itemize}
\item The introduction of EndoUFM, a novel unified framework that integrates dual foundation models to adapt large-scale vision knowledge for specialized endoscopic applications.

\item The development of an adaptive fine-tuning strategy, incorporating the novel RVLoRA and Res-DSC modules, to efficiently adapt foundation models to the unique challenges of endoscopic depth estimation. Furthermore, a foundation model-enhanced loss is designed, leveraging the pre-segmented masks generated from foundation models to enhance semantic consistency in depth predictions. 

\item Extensive experiments on four public datasets demonstrating that the proposed method achieves state-of-the-art performance and superior robustness, while maintaining an efficient model size suitable for real-world clinical deployment.
\end{itemize}

\section{Related works}
\subsection{Unsupervised monocular depth estimation based on CNN}
UMDE has seen significant advancements in natural image domains. Early work by Zhou et al. \cite{zhou2017unsupervised} pioneered the use of view synthesis for UMDE. This was further enhanced by Godard et al. \cite{godard2019digging}, who improved performance through per-pixel loss minimization and automatic masking techniques. Despite these breakthroughs in natural scenes, most of these methods rely on the assumption of photometric consistency. This reliance severely limits their effectiveness in endoscopic environments, which are characterized by highly variable illumination. To mitigate this issue, researchers have explored various strategies. For instance, Ozyoruk et al. \cite{ozyoruk2021endoslam} and Shao et al. \cite{shao2022self} first pre-processed images to adjust lighting before performing depth estimation. Another approach, proposed by Li et al. \cite{li2024image}, improved results by employing intrinsic image decomposition to separate illumination variations and then leveraging reflectance consistency. Wang et al. \cite{wang2025monopcc} introduced MonoPCC, a novel approach that fundamentally resolves brightness variations by reforming the photometric constraint into a cycle-consistent warping mechanism. Beyond these photometric consistency-based methods, Batlle et al. \cite{batlle2022photometric} utilized a photometric stereo technique for depth estimation in the human colon. Despite these advancements, existing methods often lack global scene understanding and struggle with the unique characteristics of endoscopic images.

\subsection{The application of foundation models in depth estimation}
In recent years, foundation models for depth estimation have begun to show immense potential. Compared to conventional CNNs, Vision Transformer(ViT)-based architectures of foundation models offer stronger global context modeling, effectively alleviating the limitations of local receptive fields. Ranftl et al. \cite{ranftl2020towards} introduced MiDaS to improve the model's generalization ability across different scenes by aligning the scale and shift of depth maps and mixing datasets with different annotations during training. Oquab et al. \cite{oquab2024dinov2} proposed DINOv2, a powerful visual representation foundation model that has shown outstanding performance in a wide range of computer vision tasks, including depth estimation. More recently, Depth Anything Model \cite{yang2024depth} trained a depth estimation foundation model using larger-scale labeled and unlabeled datasets and designed an auxiliary supervision to enable the model to inherit rich geometric priors from the pre-trained DINOv2 encoder. In the medical domain, the first foundation model specifically for zero-shot cross-dataset depth estimation in endoscopy was proposed by Tian et al. \cite{tian2024endoomni} with their work, EndoOmni. The model is characterized by a robust self-learning framework with a teacher-student model that learns from a mix of labeled and unlabeled data. However, due to the limited data in the medical field, it remains challenging to train and build medical-specific foundation models from scratch. Therefore, adopting general-purpose foundation models for specific task domains is a practical and effective strategy. In endoscopic depth estimation, Cui et al. \cite{cui2024surgical} first proposed Surgical-DINO, which utilizes the DINOv2 encoder for improved feature extraction. Their follow-up work, EndoDAC \cite{cui2024endodac}, applies Depth Anything Model to the depth estimation network. EndoDAC relies solely on Depth Anything Model, which struggles with complex anatomical boundaries without semantic guidance. Concurrently, Yang et al. \cite{yang2024selfCLIP} proposed an innovative framework that integrates CLIP \cite{radford2021learning} with segmentation tasks to enhance encoder performance and boost depth estimation quality. In this approach, the segmentation task serves primarily as an auxiliary loss constraint. Nevertheless, how to efficiently integrate and exploit the capabilities of different foundation models for robust and accurate endoscopic depth estimation remains an open research problem requiring further exploration. To bridge this gap, the proposed EndoUFM framework synergistically integrates dual foundation models---the Depth Anything Model and MedSAM, to effectively utilize large-scale vision knowledge for endoscopic scenes. 

\subsection{Low-rank adaptation}
With the widespread application of large models, LoRA \cite{hu2022lora} has revolutionized the efficient training of large language models for specialized tasks. Its core innovation lies in decomposing weight matrices into two low-rank components, A and B, which drastically cuts down on training resource demands. This foundational concept has since inspired numerous advancements. For instance, LoRA+ \cite{hayou2024lora+} boosts efficiency by assigning distinct learning rates to matrices $A$ and $B$. AdaLoRA \cite{zhang2023adaptive}, proposed by Zhang et al., further refines this by adaptively selecting different ranks for various adapters, thereby improving LoRA's learning capacity and training stability. VeRA \cite{kopiczko2024vera} significantly reduces trainable parameters by utilizing a single pair of frozen, randomly initialized low-rank matrices shared across all layers, learning only small scaling vectors. However, this parameter efficiency often comes at the cost of reduced accuracy. Acknowledging that the effectiveness of fine-tuning strategies varies across different domains, Cui et al. \cite{cui2024endodac} recently proposed DV-LoRA to efficiently adapt foundation models specifically for surgical scene depth estimation. DV-LoRA introduces a dynamic adaptation strategy that alternates between training low-rank matrices and scaling vectors during different training stages. Despite these advancements, existing fine-tuning methods exhibit limitations in handling the inherent monocular scale ambiguity of endoscopic scenes and often fail to capture fine-grained local details. Recognizing these gaps, there is a clear need for a fine-tuning approach specifically tailored to the unique challenges of endoscopic depth estimation. To this end, an adaptive fine-tuning strategy combining Random Vector Low-Rank Adaptation (RVLoRA) and Residual Block Based on Depthwise Separable Convolution (Res-DSC) is proposed. Unlike standard random adaptation methods (e.g., VeRA), this tailored design systematically mitigates monocular scale ambiguity through structured random perturbations while simultaneously capturing essential fine-grained local details.

\begin{figure*}
\centerline{\includegraphics[width=0.95\linewidth]{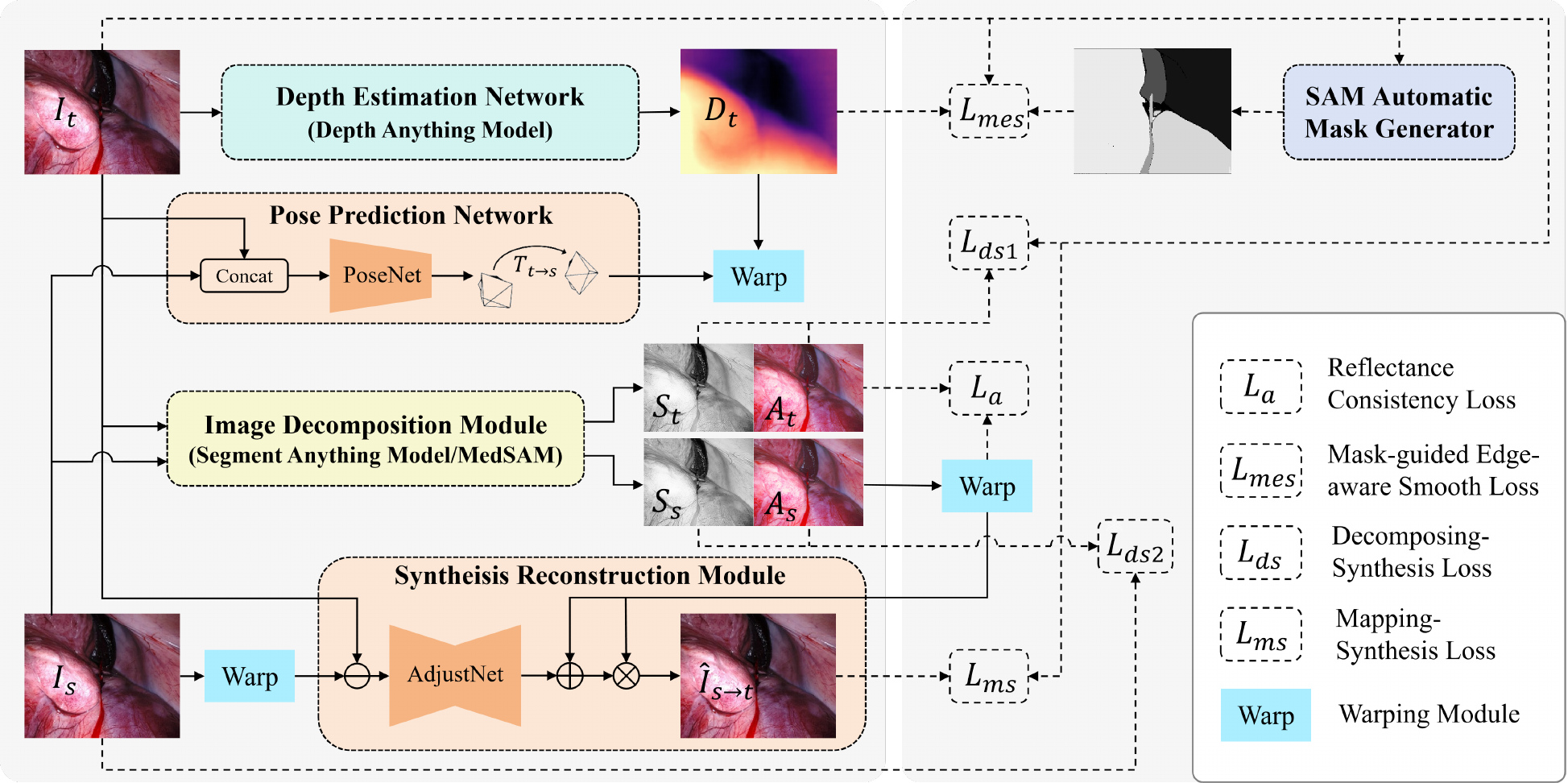}}
\caption{Network architecture. The depth estimation network is built on the Depth Anything Model and the image decomposition network is augmented with the MedSAM.}
\label{fig2}
\end{figure*}

\section{Method}
In this section, the architecture of EndoUFM, a novel framework for unsupervised monocular depth estimation, is presented. The integration of foundation models, the adaptive fine-tuning strategy, and the custom loss functions that collectively enable state-of-the-art performance on endoscopic images are detailed.

\begin{figure*}
\centerline{\includegraphics[width=0.95\linewidth]{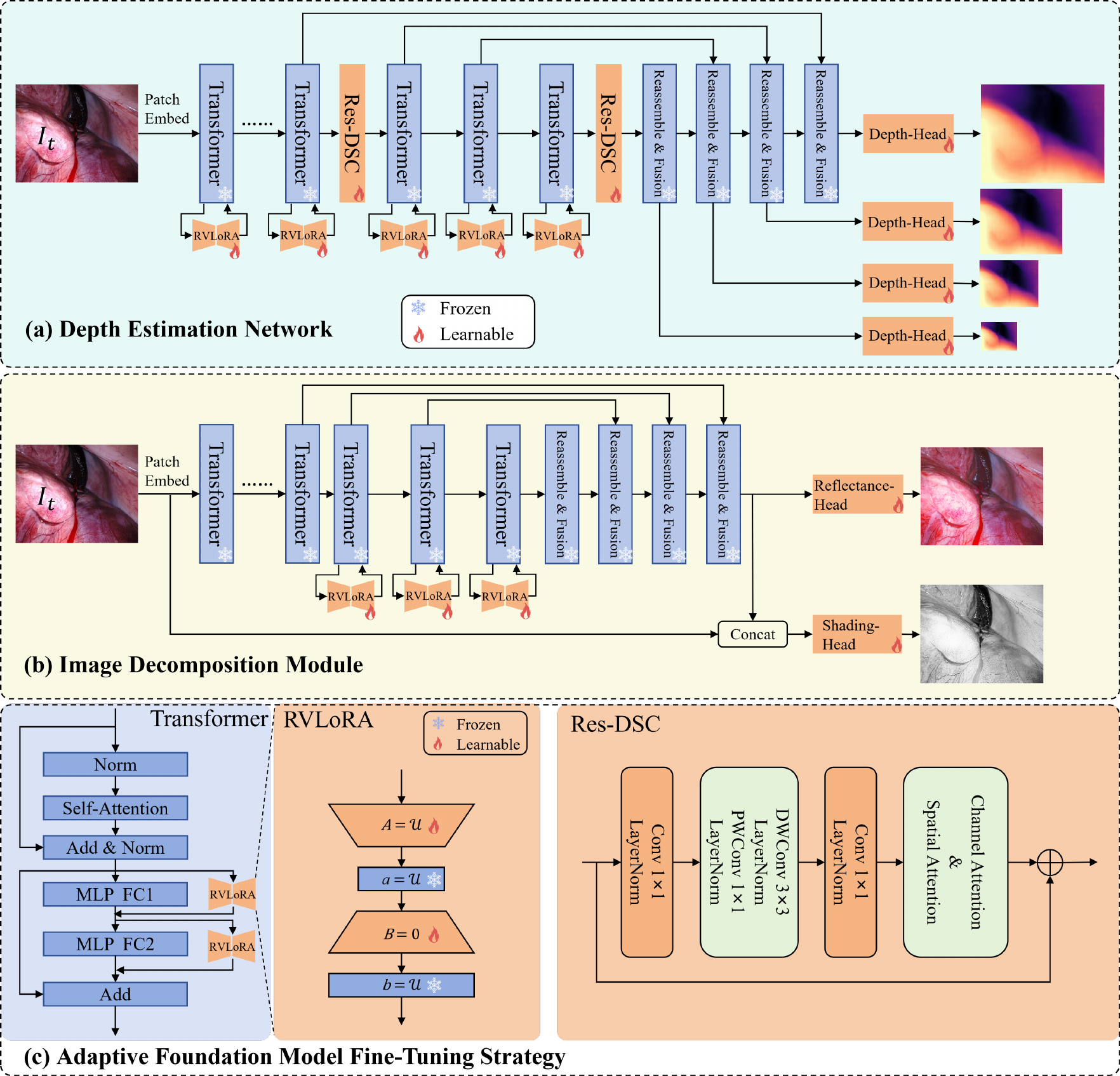}}
\caption{(a) Depth estimation network. (b) Image Decomposition Module. (c) Adaptive Foundation Model Fine-Tuning Strategy. left: The deployment and detailed structure of RVLoRA. RVLoRA is connected to the two linear layers of the MLP in each transformer module. The parameters $a$ and $b$ are frozen while $A$ and $B$ are trainable. right: The detailed structure of Res-DSC. Res-DSC contains a depthwise separable convolution, a pointwise separable convolution, a channel attention module, a spatial attention module, and the residual connection.}
\label{fig3}
\end{figure*}

\subsection{EndoUFM framework integrated with foundation models}

A novel decomposition-based unsupervised monocular depth estimation framework is proposed, which is designed to enhance robustness and accuracy in endoscopic scenarios by leveraging foundation models. The improvement consists of two main components: a depth estimation network built on the Depth Anything Model and an image decomposition network augmented with the Segment Anything Model, specifically MedSAM. These components are designed to address the domain shift and semantic inconsistency challenges prevalent in endoscopic images. 

The proposed network, as illustrated in Fig. \ref{fig2}, is based on the previous work IID-SfMLearner \cite{li2024image}.The target frame is defined as $I_t$ and the adjacent source frames are denoted as $I_s$. The network comprises a depth estimation network $\Phi_D$ for predicting single-frame depth, a pose estimation network $\Phi_T$ to estimate the relative inter-frame camera pose $T_{t\rightarrow s}$, a decomposition module $\Phi_I$ to decompose images into reflectance ($R$) and shading ($S$) maps, and a synthesis and reconstruction module $\Phi_L$ to reconstruct the target frame $\hat{I}_{s\rightarrow t}$ from the decomposed components.

Classical monocular depth estimation methods typically employ standard convolutional neural networks for depth prediction, which often lack sufficient global perception to accurately capture depth information. To improve depth prediction, the conventional network is replaced with the Depth Anything Model transformer backbone, enabling the model to capture long-range geometric priors while retaining local anatomical details. As shown in the Fig. \ref{fig3}(a), the encoder and decoder maintain the same structure as the Depth Anything Model. Similar to the previous work \cite{cui2024endodac,peng2025hadepth}, features from the last four encoder layers are fed into its frozen Reassemble \& Fusion module, which generates and aggregates features at different resolutions. The features then pass through the trainable Depth-Head module for final dense prediction, yielding depth estimation outputs at four scales.

Furthermore, the powerful architecture of the MedSAM model \cite{kirillov2023segment, ma2024segment} is integrated into the image decomposition network. This infusion of semantic information significantly enhances the network's awareness and decomposition capabilities in scenes with complex textures and varying illumination, overcoming the limitations of previous ResNet-based decomposition networks \cite{li2024image}. As shown in Fig. \ref{fig3}(b), the encoder is same with the MedSAM’s vision transformer architecture and the decoder structure is similar to that of the depth estimation network, with the Depth-Head modified to a Reflectance-Head and a Shading-Head for image decomposition. Features extracted by the MedSAM encoder are firstly decoded to generate the reflectance map. Then the shallow features from the decoder are concatenated with the original image to incorporate fine details, and then processed by the shading head to predict the shading map.

\subsection{Adaptive foundation model fine-tuning strategy}
Fine-tuning foundation models presents a direct approach to enhance long-range dependencies without significantly increasing the number of trainable parameters. However, adapting a model like Depth Anything Model, pre-trained on natural images, to the unique characteristics of endoscopic data requires a sophisticated strategy. Standard fine-tuning methods can lack robustness, particularly in addressing the inherent scale ambiguity of monocular depth estimation and capturing fine-grained local details \cite{Bian2019unsupervisedscale,Song2024unsupervisedconsistency,ZHANG2026Multi}. To overcome these challenges, a novel adaptive fine-tuning strategy is proposed, as shown in Fig. \ref{fig3}(c), which synergistically integrates Low-Rank Adaptation based on Random Vectors (RVLoRA) and Residual blocks based on Depthwise Separable Convolution (Res-DSC).

\subsubsection{Low-Rank Adaptation based on Random Vectors (RVLoRA)}
While neural networks often contain numerous dense layers with full-rank weight matrices, pre-trained foundation models demonstrate a surprisingly low "intrinsic dimension" when adapted to specific tasks \cite{aghajanyan2021intrinsic}. This suggests that weight updates can be effectively constrained to a low-dimensional subspace. Inspired by this, Low-Rank Adaptation (LoRA) was proposed \cite{hu2022lora}, positing that weight updates during adaptation also exhibit a low "intrinsic rank". For a pre-trained weight matrix $W_0\in\mathbb{R}^{m\times n}$, its update can be efficiently represented through low-rank decomposition as $W_0+\Delta W=W_0+BA$. The modified forward pass then becomes:
\begin{equation}
h=W_{0}x+\Delta Wx=W_{0}x+BAx 
\tag{1}
\end{equation}
where $B\in\mathbb{R}^{m\times r}$ and $A\in\mathbb{R}^{r\times n}$, the rank $r$ is significantly smaller than both $m$ and $n$ ($r \ll \min (m, n)$). During the training process, $W_0$ remains frozen and does not receive gradient updates. Instead, only the low-rank matrices $A$ and $B$ are trainable, drastically reducing the number of trainable parameters for downstream tasks while keeping the core pre-trained model weights intact.

Recent research has revealed the potent role of fixed, random parameters in deep learning frameworks, offering a powerful solution to one of the most significant challenges in monocular depth estimation: scale ambiguity. This inherent challenge means models struggle to perceive actual scene scale, demanding exceptional robustness to scale variations. As the studies have shown that introducing random vectors into LoRA can significantly boost a model's adaptive capacity \cite{kopiczko2024vera, lu2022frozen}, this can be an efficient mechanism for fine-tuning a model to mitigate the challenge of scale ambiguity. Moreover, within the affine transformations heavily leveraged by deep learning, freezing randomly initialized parameters is a critical technique. It enhances the network's ability to capture and generalize these transformations by preventing the model from overfitting to specific transform patterns seen during training \cite{frankle2021training}, thereby strengthening the robustness of the low-rank matrices $A$ and $B$.

Building on these insights, a parameter-efficient fine-tuning method is proposed: Low-Rank Adaptation based on Random Vectors (RVLoRA). The proposed method incorporates frozen, randomly initialized vectors a and b into the LoRA framework. These vectors provide a continuous and stable random perturbation to the model's internal weights, creating a multiple set of scale transformations, and compelling the model to learn essential and generalizable geometric relationships rather than the absolute depth of specific scenes. By preventing overfitting to the training data's scale, the frozen random vectors make the model fundamentally robustly adapt to scale changes, enabling it to deliver consistent and accurate depth estimations across diverse new scenes.

During the training phase, Kaiming uniform initialization \cite{he2015delving} is employed for vectors $a$, $b$, and the low-rank matrix $A$, while the low-rank matrix $B$ is initialized to zeros. This specific initialization strategy ensures that the original weight matrix remains unaffected during the first forward pass. The frozen scaling vectors and the trainable low-rank matrices are combined with the original, pre-trained weights without causing additional inference delay. The proposed method can be formally represented as:
\begin{equation}
h=W_0 x+\Delta W x=W_0 x+\Lambda_b B \Lambda_a A x
\tag{2}
\end{equation}
where $B\in\mathbb{R}^{m\times r}$ and $A\in\mathbb{R}^{r\times n}$ represent the trainable low-rank matrices, and the rank $r$ is much smaller than both $m$ and $n$ ($r \ll \min (m, n)$). The novel components are $\Lambda_b\in\mathbb{R}^{m\times m}$ and $\Lambda_a\in\mathbb{R}^{r\times r}$. These are diagonal matrices derived from the randomly initialized vectors $b\in\mathbb{R}^{m\times 1}$ and $a\in\mathbb{R}^{r\times 1}$, respectively. Crucially, $b$ and $a$ are randomly initialized and then remain frozen throughout the entire training process. This approach effectively applies random scaling to the rows of the low-rank matrices $B$ and $A$, which in turn enhances the model's adaptability across layers of the network.

As shown in Fig. \ref{fig3}(a), the pre-trained Depth Anything Model is applied and all the transformer blocks in the encoder are fine-tuned to specifically enhance the accuracy of depth estimation. The MedSAM encoder framework employed is detailed in Fig. \ref{fig3}(b). To adapt to endoscopic images, a pre-trained MedSAM model \cite{ma2024segment} is utilized allowing the model to leverage learned medical prior information. To optimize training parameters and time, only the bottom three transformer blocks of the encoder are fine-tuned \cite{shvets2024joint}, which demonstrably improves performance.

\subsubsection{Residual block based on Depthwise Separable Convolution (Res-DSC)}
While transformers excel at dynamic attention, global context awareness, and generalization, they often struggle with extracting fine details and local features. Conversely, convolutions are adept at capturing local features and offer inherent benefits such as shift, scale, and distortion invariance. Combining CNNs with Transformers can create a synergistic effect, significantly boosting overall model performance. 

Inspired by this, a Residual block based on Depthwise Separable Convolution (Res-DSC) is designed, as shown in Fig. \ref{fig3}(c). This structure utilizes three convolutional layers: an initial $1 \times 1$ convolution layer for channel reduction, a $3 \times 3$ depth-wise separable convolution, which reduces parameters for faster computation and a more streamlined model compared to traditional convolutions. A $1 \times 1$ convolution then restores the number of channels. Channel attention and spatial attention modules are also incorporated \cite{woo2018cbam}. By combining these, the input features are doubly refined, allowing the model to simultaneously focus on which channels and spatial locations of the features are the most important. 

To minimize model parameters while maintaining depth estimation accuracy, Res-DSC blocks are incorporated only after the 3rd, 6th, 9th, and 12th transformer blocks of the depth estimation network. For brevity, Fig. \ref{fig3}(a) only illustrates the Res-DSC blocks after the 9th and 12th transformer blocks.

In summary, the roles of the two modules in adaptive fine-tuning strategy can be summarized as follows: 
\begin{itemize}
\item RVLoRA addresses scale ambiguity and overfitting by introducing stable random perturbations in the low-rank adaptation feature space, encouraging scale-robust representation learning.

\item Res-DSC addresses insufficient local features sensitivity in ViT-based encoders by injecting lightweight convolutional inductive bias into the depth estimation branch.
\end{itemize}

RVLoRA and Res-DSC are complementary rather than redundant. The former regularizes how the model adapts, while the latter improves what features are emphasized for depth estimation.

\subsection{Foundation model-enhanced loss design}
Similar to the loss function design of IID-SfMLearner, a synthetic frame $I_{s\rightarrow t}$ is first reconstructed via a warping transformation using the predicted depth $D_t$ from the depth estimation network and the relative pose $T_{t\rightarrow s}$. Based on the image decomposition theory \cite{krebs2020intrinsic}, a reflectance consistency loss $L_a$ is formulated to minimize the reflectance difference between adjacent frames. A decomposing-synthesis loss $L_{ds}$ and a mapping-synthesis loss $L_{ms}$ are also introduced to constrain the decomposition and reconstruction processes. The expressions for these three loss functions are as follows:
\begin{equation}
L_{a}(R_{t}, R_{s \to t}) = \|R_{t} - R_{s \to t}\|_{1}
\tag{3}
\end{equation}
\begin{equation}
\left\{
\begin{aligned}
L_{ds1}(\hat{I}_{t}, I_{t}) &= \alpha \frac{1 - SSIM(\hat{I}_{t}, I_{t})}{2} + (1-\alpha)\|\hat{I}_{t} - I_{t}\|_{1} \\
L_{ds2}(\hat{I}_{s}, I_{s}) &= \alpha \frac{1 - SSIM(\hat{I}_{s}, I_{s})}{2} + (1-\alpha)\|\hat{I}_{s} - I_{s}\|_{1}
\end{aligned}
\right.
\tag{4}
\end{equation}
\begin{equation}
L_{ms}(\hat{I}_{s \to t}, I_{t}) = \alpha \frac{1 - SSIM(\hat{I}_{s \to t}, I_{t})}{2} + (1-\alpha)\|\hat{I}_{s \to t} - I_{t}\|_{1}
\tag{5}
\end{equation}
where $\alpha$ is a hyperparameter to balance the L1 loss and SSIM loss, and set to 0.85 in the proposed method. 

To further leverage the profound semantic segmentation power of the MedSAM model, automatically generated segmentation masks from the pre-trained model without providing any prompt are utilized to construct a mask-guided depth smoothness loss. This loss promotes depth consistency within each semantic region. The specific loss function is expressed as: 
\begin{equation}
L_{mes}(D_{t}, I_{t}) = \sum M_{i} \left( |\partial_{x} D_{t}| e^{-|\partial_{x} I_{t}|} + |\partial_{y} D_{t}| e^{-|\partial_{y} I_{t}|} \right)
\tag{6}
\end{equation}
where $M_{i}$ represents the auto-generated masks, $\partial D_{t}$ and $\partial I_{t}$ are the gradients of the depth map and image, respectively. This loss acts as an auxiliary regularizer to refine the depth estimation. Its key advantage is that the segmentation mask prevents incorrect smoothing between semantically different regions, directly channeling the foundation model's advanced perceptual capabilities to mitigate ambiguity and refine the final depth predictions. 

Integrating the various modules discussed above, the final loss function is defined as: 
\begin{equation}
loss = \lambda_{ds} L_{ds} + \lambda_{a} L_{a} + \lambda_{ms} L_{ms} + \lambda_{mes} L_{mes}
\tag{7}
\end{equation}
where $L_{ds}$, $L_{a}$, $L_{ms}$, and $L_{mes}$ are weighting factors. All modules are trained end-to-end using self-supervised objectives, including decomposing-synthesis loss, reflectance consistency loss, mapping-synthesis loss, and the proposed mask-guided smoothness loss.

\begin{table}[!t]
  \centering
    \caption{The calculation of applied evaluation metrics, in which $d$ represents the prediction depth and $d^*$ represents the ground truth.}
    \label{tab1}
    \resizebox{0.5\textwidth}{!}{
    \begin{tabular}{c | c} 
    \hline
    \textbf{Metric}	&\textbf{Definition}\\
    \hline 
       Abs Rel & $\frac{1}{|N|}\sum_{d\in N}|d-d^*|/d^*$\\
      Sq Rel & $\frac{1}{|N|}\sum_{d\in N}|d-d^*|^2/d^*$\\
      RMSE & $\sqrt{\frac{1}{|N|}\sum_{d\in N}(d-d^*)^2}$\\
      RMSE log & $\sqrt{\frac{1}{|N|}\sum_{d\in N}(\ln d-\ln d^{*})^{2}}$\\
      $\delta1$, $\delta2$, and $\delta3$  & $\frac{1}{|N|} \sum [ \max \left(\frac{d}{d^*}, \frac{d^*}{d}\right)<T ], T\in \{1.25, 1.25^2, 1.25^3\} $\\
    \hline
    \end{tabular}
    }
\end{table}

\section{Experiments}
\subsection{Datasets and metrics}
Our experiments are conducted on four publicly available datasets: SCARED \cite{allan2021stereo}, Hamlyn \cite{mountney2010three}, SERV-CT \cite{edwards2022serv}, and EndoNeRF \cite{wang2022neural}. Training and test are performed on the SCARED and Hamlyn datasets, and the trained network is directly applied on the SCARED to the SERV-CT and EndoNeRF datasets to further validate the advantage of the proposed method.

SCARED dataset, originating from the 2019 MICCAI challenge, comprises nine subsets collected from porcine cadavers using the da Vinci surgical robot. Each subset contains 4 to 5 stereo video sequences, with ground truth depth maps obtained using structured light encoding during the acquisition process. For monocular depth estimation, as in previous works \cite{li2024image, shao2022self}, only the left-view images are used, which are uniformly resized to $320\times256$. Following AF-SfMLearner \cite{shao2022self}, the dataset is divided into 15,351 frames for training, 1,705 frames for validation, and 551 frames for testing. 

Hamlyn dataset is an in vivo endoscopy stereo video dataset provided by the Hamlyn Center for Laparoscopy at Imperial College London. The rectified version provided by Recasens et al. \cite{recasens2021endo}, is adopted, along with the ground truth depth maps for left-view frames. The image size is $288\times256$, with 12,796 frames allocated for training, 2,076 for validation, and 2,319 for testing.

SERV-CT dataset consists of 16 endoscopic stereo image pairs captured using the da Vinci surgical robot on two different ex vivo porcine samples. Depth maps for the endoscopic images were computed by precisely aligning 3D models derived from CT images with the endoscopic stereo images. The test dataset contains a total of 32 frames, each with a resolution of $720\times576$. 

EndoNeRF dataset includes two endoscopic scenes generated by the da Vinci surgical robot, depicting "cutting tissues twice" and "pulling soft tissues." Each image has a resolution of $640\times512$ and is accompanied by corresponding depth maps and surgical tool masks. The stereo depth maps are obtained using STTR-Light \cite{li2021revisiting}, and the tool masks are extracted manually. As with the previous datasets, 219 left-view images are used for testing. 

Seven commonly used metrics are employed in the evaluation, i.e., Abs Rel, Sq Rel, RMSE, RMSE log, and Threshold $\delta1$, $\delta2$, $\delta3$, which are listed in Table \ref{tab1}. Same with other methods \cite{zhou2017unsupervised, shao2022self}, the predicted depth map is scaled due to its unknown scale. The scaling factor is the ground truth median depth divided by the predicted median depth, which can be expressed as:
\begin{equation}
  f_{scale}=median(d^{*})/median(d)
\tag{8}
\end{equation}
where $d$ represents the prediction depth, and $d^{*}$ represents the ground truth. The scaled depth map is capped at 150 mm for all the datasets, which can cover all pixels.

To evaluate camera pose estimation performance, the predicted camera trajectory is aligned to the scale of the ground-truth trajectory and the Absolute Trajectory Error (ATE) metric computed over 5-frame segments is employed as the evaluation metric.

\subsection{Implementation details}
All the models are trained end-to-end using PyTorch \cite{paszke2019pytorch} and the Adam optimizer \cite{Kingma2014AdamAM} with $\beta_1 = 0.9$ and $\beta_2 = 0.999$. The proposed model is trained for 30 epochs on a GeForce A100 GPU with a batch size of 8. For training on the SCARED dataset, the initial learning rate is set to $1 \times 10^{-4}$ and is multiplied by a scale factor of 0.1 after 10 epochs. In the experiments, the rank for RVLoRA is set to 4, the loss weights $L_{ds}$, $L_{a}$, $L_{ms}$ and $L_{mes}$ are set to 0.2, 0.2, 1, and 0.003, respectively. When training on the Hamlyn dataset, the initial learning rate is adjusted to $5 \times 10^{-5}$, and $L_{mes}$ is set to 0.006. All other hyperparameters remain consistent with the SCARED dataset training.

\begin{table*}[t]
  \centering
    \caption{Comparative experimental results on the SCARED dataset and parameter counts of different methods. The best results are presented in bold and the second-best underlined.}
    \label{tabscared}
    \resizebox{1.0\textwidth}{!}{
    \begin{tabular}{c|c c c c c c c|c c } 
    \hline
      Method&Abs Rel$\downarrow$&Sq Rel$\downarrow$&RMSE$\downarrow$&RMSE log$\downarrow$&$\delta1 \uparrow$&$\delta2 \uparrow$&$\delta3 \uparrow$&Total.(M)&Train.(M)\\
        \hline 
       Monodepth2 \cite{godard2019digging} &   0.065  &   0.560  &   5.682  &   0.092  &   0.959 &   0.996  & \underline{0.999} & 14.84 & 14.84\\
       AF-SfMLearner \cite{shao2022self} &   0.060  &   0.443  &   4.964  &   0.082  &   0.973  &   \underline{0.998}  & \textbf{1.000} & 14.84 & 14.84\\
        IID-SfMLearner \cite{li2024image} &   0.058  &   0.430  &   4.808  &   0.080  &   0.972  &   \underline{0.998}  & \underline{0.999} & 14.84 & 14.84\\ 
        MonoPCC \cite{wang2025monopcc} & 0.053  & 0.368  & 4.616  & 0.075 & \underline{0.981} & \underline{0.998} & \textbf{1.000} & - & -\\
        \hline
         Depth Anything Model \cite{yang2024depth} & 0.086  & 0.927  & 6.956  & 0.112  & 0.929 & 0.993 & \underline{0.999} & 97.50 & 97.50\\ 
         EndoDAC \cite{cui2024endodac} &   \underline{0.051}  &   \underline{0.355}  &   \underline{4.442}  & \underline{0.073}  &   0.979  &   \underline{0.998}  &   \textbf{1.000} & 99.09 & 1.66\\ 
        Ours  & \textbf{0.050}  & \textbf{0.317}  & \textbf{4.141}  & \textbf{0.070}  & \textbf{0.982} &   \textbf{0.999}  &  \textbf{1.000}& 99.10 & 1.67\\
        \hline
     \end{tabular}
    } 
   
\end{table*}

\begin{table*}[t]
  \centering
    \caption{Comparative experimental results on the Hamlyn dataset. The best results are presented in bold and the second-best underlined.}
    \label{tabhamlyn}
    \resizebox{0.8\textwidth}{!}{
    \begin{tabular}{c|c c c c c c c} 
    \hline
      Method&Abs Rel$\downarrow$&Sq Rel$\downarrow$&RMSE$\downarrow$&RMSE log$\downarrow$&$\delta1 \uparrow$&$\delta2 \uparrow$&$\delta3 \uparrow$ \\
    \hline
    Monodepth2 \cite{godard2019digging} &   0.094  &   0.718  &   5.566  &   0.118  &   0.921  & \underline{0.997}  & \textbf{0.999}\\
    AF-SfMLearner \cite{shao2022self} &   0.087  &   0.656  &   5.134  &   0.108  &   0.925  &   \textbf{0.998}  &   \textbf{0.999}\\
    IID-SfMLearner \cite{li2024image} &   0.085  &   0.619  &   5.055  &   \underline{0.106}  &   0.936  &   \textbf{0.998}  &   \textbf{0.999} \\
    MonoPCC \cite{wang2025monopcc}  &  \underline{0.083}  & \underline{0.550}  &   \underline{4.942}  &   \underline{0.106}  &  \underline{0.959}  & \underline{0.997}  &   \textbf{0.999} \\
    \hline
    Depth Anything Model \cite{yang2024depth} & 0.099 & 0.979 & 6.193 & 0.125 & 0.901 & 0.989 & \underline{0.998} \\
    EndoDAC \cite{cui2024endodac} &   0.088  &   0.666  &   5.203  &   0.109  &   0.929  &   \textbf{0.998}  & \textbf{0.999} \\
    Ours      &   \textbf{0.071}  &   \textbf{0.445}  &   \textbf{4.280}  &   \textbf{0.089}  &   \textbf{0.973}  &   \textbf{0.998}  &   \textbf{0.999}\\
     \hline
     \end{tabular}
    } 
   
\end{table*}

\subsection{Comparison experiments}
The proposed method was compared with six leading unsupervised monocular depth estimation methods, i.e., Monodepth2 \cite{godard2019digging}, AF-SfMLearner \cite{shao2022self}, IID-SfMLearner \cite{li2024image}, Depth Anything Model \cite{yang2024depth}, EndoDAC \cite{cui2024endodac}, and MonoPCC \cite{wang2025monopcc}. For a fair comparison, all methods were evaluated using the same input image resolution and an identical data preprocessing pipeline. The methods Monodepth2 and IID-SfMLearner were retrained using their publicly released codes, while the other methods were tested using the trained model provided by the authors. All compared methods share an identical training dataset configuration and their performance was compared on the same test dataset. The average results over three independent runs were reported to ensure a more stable and reliable evaluation. 

The quantitative results on the SCARED dataset, detailed in Table \ref{tabscared}, demonstrate the superiority of the EndoUFM framework. The proposed model achieves state-of-the-art performance across all seven evaluation metrics. Earlier methods like Monodepth2, designed for natural images, struggle with the unique challenges of endoscopic scenes, such as variable lighting and specular reflections. While AF-SfMLearner and IID-SfMLearner introduce mechanisms to handle illumination inconsistencies, and MonoPCC utilize frequency domain information to achieve better results, the proposed method surpasses them all. The direct application of the Depth Anything Model, pre-trained on natural images, yields suboptimal results due to the significant domain gap. EndoDAC, which represents an initial effort in fine-tuning foundation models for endoscopy, shows improved performance. However, the proposed method, with its more sophisticated integration of dual foundation models and advanced fine-tuning strategy, sets a new benchmark. Notably, EndoUFM reduces the Sq Rel by 13.86\% compared to MonoPCC and 10.70\% compared to EndoDAC, and it lowers the RMSE by 10.29\% and 6.78\% against the same models, respectively.

In terms of model complexity, the proposed model has only 1.67M trainable parameters, which is merely 1.7\% of the parameters of the depth estimation network. This indicates that the method achieves an excellent balance between lightweight design and performance, making it highly suitable for practical deployment in minimally invasive surgical robot systems. 

\begin{figure*}
\centerline{\includegraphics[width=0.9\linewidth]{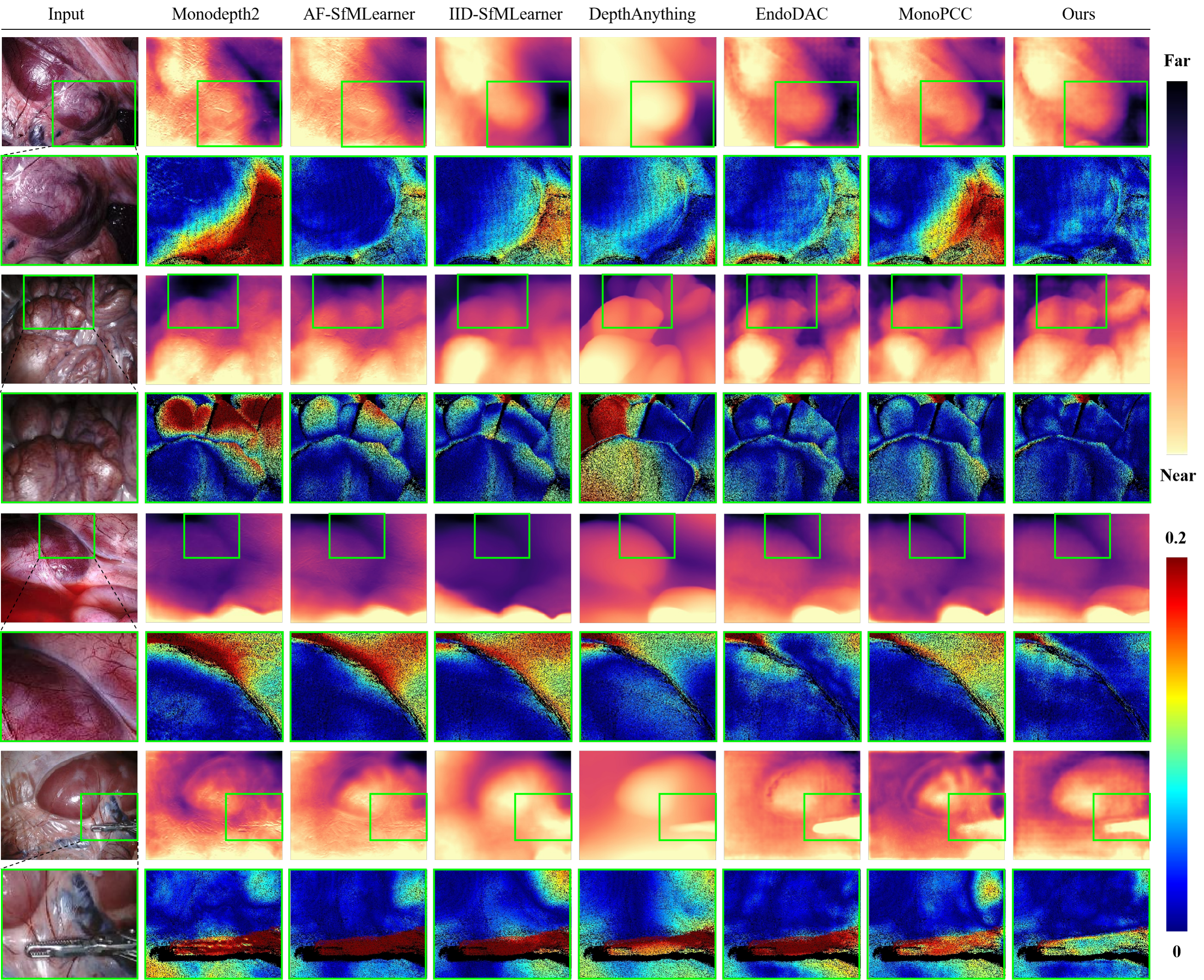}}
\caption{Qualitative comparison of depth estimation and Abs Rel error maps on the SCARED dataset. The left column shows the input image. Subsequent columns display the depth prediction and the corresponding error map for each method. The regions of interest (ROIs) are highlighted with green boxes. The color scale for the error maps ranges from blue (small error) to red (large error), with a maximum error value of 0.2.}
\label{fig4}
\end{figure*}

Fig. \ref{fig4} displays the qualitative depth estimation results of the comparative methods on the SCARED dataset, highlighting the noticeable differences across various approaches. For a clearer visual comparison, error maps are generated to show the Absolute Relative (Abs Rel) error. This technique, following the methodology of Wang et al. \cite{wang2025monopcc}, translates the pixel-level error into a color-coded map for easier interpretation. For each image, the highlighted region is magnified to better display the visualized error maps of the different methods. Clearly, our method produces smaller errors compared to other approaches, particularly along tissue boundaries and in low-light areas. As seen in the magnified areas, our model more accurately estimates depth in scenarios with complex tissue structures and specular reflections, showcasing its robustness in challenging conditions.

\begin{figure*}
\centerline{\includegraphics[width=0.9\linewidth]{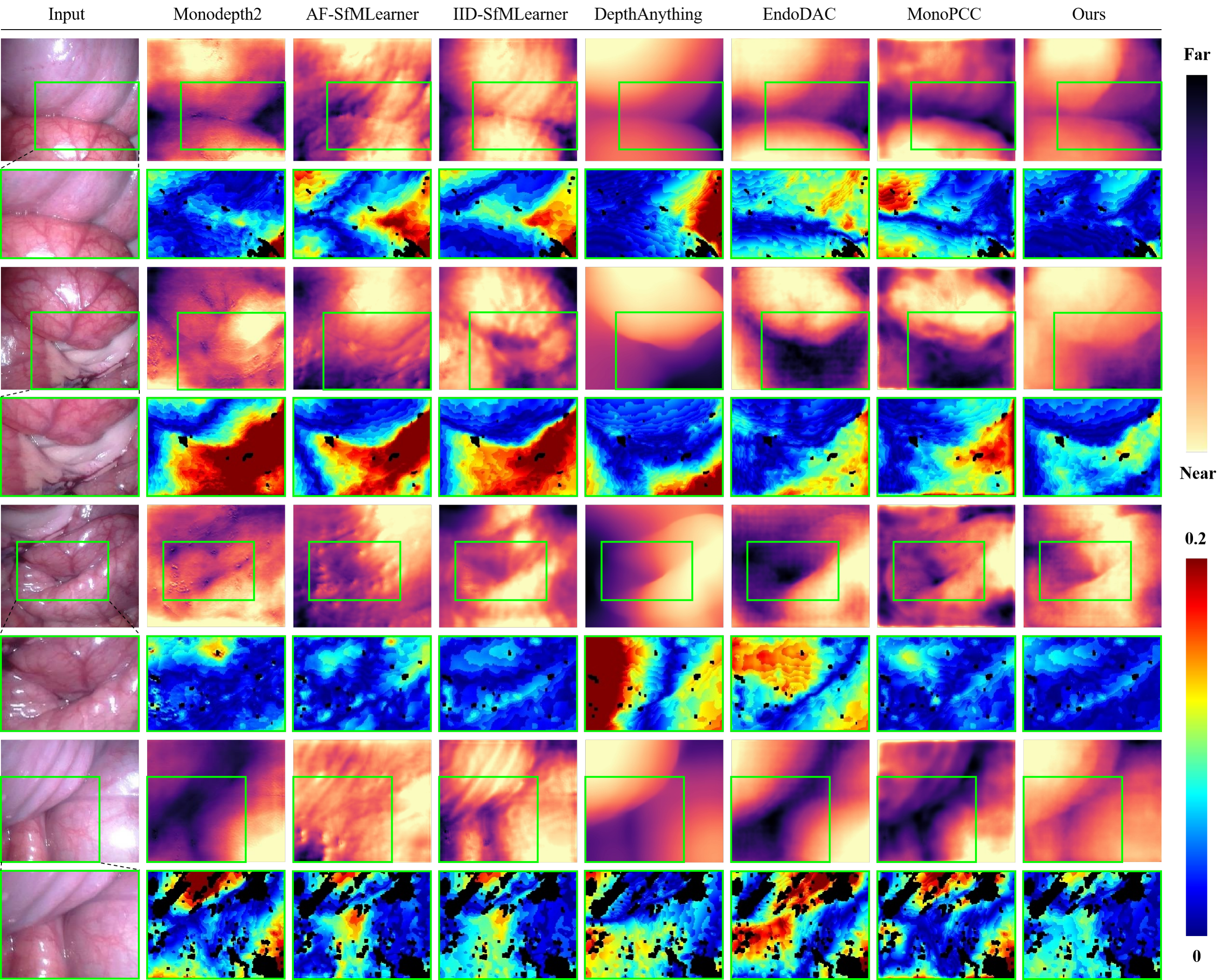}}
\caption{Qualitative comparison of depth estimation and Abs Rel error maps on the challenging Hamlyn dataset. The color scale and ROI notation are consistent with Fig. \ref{fig4}.}
\label{fig5}
\end{figure*}

EndoUFM was further validated on the Hamlyn dataset, which is characterized by lower resolution, more complex lighting conditions, and greater camera motion. As shown in Table \ref{tabhamlyn}, the proposed method again outperforms all competing approaches significantly. The qualitative results in Fig. \ref{fig5} demonstrate that the proposed model successfully preserves sharp edges even where tissue boundaries are ambiguous and produces smooth, accurate depth maps for homogeneous tissue areas.

To evaluate the generalization capabilities, the model trained on the SCARED dataset was directly applied to the SERV-CT and EndoNeRF datasets without any fine-tuning. The quantitative results, presented in Table \ref{tabservct} and \ref{tabendonerf}, highlight the robustness and effectiveness of our approach. As shown in Table \ref{tabservct}, on the SERV-CT dataset, our method establishes new state-of-the-art results across all evaluation metrics. Notably, the performance margin over the second-best method is substantial, underscoring the clear superiority of our model in this domain. On the more challenging EndoNeRF dataset, our method remains highly competitive, achieving either the best or second-best performance on all metrics. While securing the top results in four metrics, including Sq Rel and RMSE, it performs comparably to the leading method on others. For the metrics where our method is second-best, the difference is marginal, indicating that our method's performance closely approaches the optimal on this dataset.

This slight variation in performance between the two datasets is expected in a direct cross-dataset evaluation, as it reflects the inherent domain shift and differing image characteristics of each environment. Overall, these results strongly validate the excellent generalization ability of our model.

\begin{table*}[t]
  \centering
    \caption{Comparative experimental results on the SERV-CT dataset. The best results are presented in bold and the second-best underlined.}
    \label{tabservct}
    \resizebox{0.8\textwidth}{!}{
    \begin{tabular}{c|c c c c c c c} 
    \hline
      Method&Abs Rel$\downarrow$&Sq Rel$\downarrow$&RMSE$\downarrow$&RMSE log$\downarrow$&$\delta1 \uparrow$&$\delta2 \uparrow$&$\delta3 \uparrow$ \\
        \hline
        Monodepth2 \cite{godard2019digging} & 0.149 & 3.808 & 17.361 & 0.200 & 0.736 & 0.936 & 0.993 \\
        AF-SfMLearner \cite{shao2022self} & 0.124 & 2.233 & 13.125 & 0.155 & 0.845 & 0.985 & \textbf{1.000} \\
        IID-SfMLearner \cite{li2024image} & 0.138 & 1.762 & \underline{9.054} & 0.166 & 0.805 & 0.988 & \textbf{1.000} \\
        MonoPCC \cite{wang2025monopcc} & 0.089 & 1.141 & 9.527 & 0.113 & 0.927 & \textbf{0.999} & \textbf{1.000} \\
        \hline
        Depth Anything Model \cite{yang2024depth} & 0.125 & 2.984 & 13.163 & 0.145 & 0.850 & 0.970 & \underline{0.995}  \\
        EndoDAC \cite{cui2024endodac} & \underline{0.082} & \underline{1.088} & 9.208 & \underline{0.109} & \underline{0.934} & \underline{0.998} & \textbf{1.000} \\
        Ours      &\textbf{0.072} &\textbf{0.826} & \textbf{8.098} & \textbf{0.095} & \textbf{0.962} & \textbf{0.999}  & \textbf{1.000} \\
     \hline
     \end{tabular}
    } 
   
\end{table*}
\begin{table*}[t]
  \centering
    \caption{Comparative experimental results on the EndoNeRF dataset. The best results are presented in bold and the second-best underlined.}
    \label{tabendonerf}
    \resizebox{0.8\textwidth}{!}{
    \begin{tabular}{c|c c c c c c c} 
    \hline
      Method&Abs Rel$\downarrow$&Sq Rel$\downarrow$&RMSE$\downarrow$&RMSE log$\downarrow$&$\delta1 \uparrow$&$\delta2 \uparrow$&$\delta3 \uparrow$ \\
    \hline
    Monodepth2 \cite{godard2019digging} & 0.223 & 5.484 & 19.417 & 0.278 & 0.543 & 0.885 & 0.993\\
    AF-SfMLearner \cite{shao2022self} & 0.249 & 6.898 & 21.388 & 0.308 & 0.571 & 0.815 & 0.971\\
    IID-SfMLearner \cite{li2024image} & 0.225 & 5.760 & 19.958 & 0.286 & 0.542 & 0.881 & 0.994 \\
    MonoPCC \cite{wang2025monopcc}  & 0.210 & 5.070 & 19.250 & 0.274 & 0.547 & 0.891 & 0.995 \\
    \hline
    Depth Anything Model \cite{yang2024depth} & 0.230 & 5.669 & 21.061 & 0.266 & 0.504 & 0.923 & 0.995 \\
    EndoDAC \cite{cui2024endodac}& \textbf{0.161} & \underline{3.017} & \underline{16.061} & \textbf{0.203} & \underline{0.725} & \underline{0.974} & \textbf{0.998} \\
    Ours      & \underline{0.163} & \textbf{2.926} & \textbf{15.818} & \underline{0.207} & \textbf{0.728} & \textbf{0.976} & \underline{0.997}\\
     \hline
     \end{tabular}
    } 
   
\end{table*}

\subsection{Ablation study}
To further demonstrate the effectiveness of our proposed modules, relevant ablation studies were conducted on the SCARED dataset. 

The ablation study in Table \ref{tabablation} systematically evaluates the contribution of each component within the EndoUFM framework. When the pre-trained Depth Anything Model is introduced, the error rates remain relatively high. Fine-tuning this model with RVLoRA significantly improves performance, confirming the necessity of domain adaptation. The addition of the Res-DSC module further boosts performance, validating its effectiveness in capturing fine-grained local features. 

Similar improvements can be observed when incorporating the pre-trained MedSAM and fine-tuning it with RVLoRA in the image decomposition network, highlighting the value of its powerful semantic feature extraction. However, adding the Res-DSC module to the MedSAM results in a decline in performance. Therefore, it is not included in the final model. Finally, the complete EndoUFM model, incorporating the mask-guided smoothness loss, achieves the best performance across all metrics. This demonstrates that every component in EndoUFM is vital for improving overall accuracy and robustness.

\begin{table*}[t]
  \centering
    \caption{The ablation study results of the proposed modules. The best results are presented in bold.}
    \label{tabablation}
    \resizebox{1.00\textwidth}{!}{
    \begin{tabular}{c c c | c c c | c | c c c c c c c} 
    \hline
    \multicolumn{3}{c|}{\textbf{DepthAnything}} & \multicolumn{3}{c|}{\textbf{SegmentAnything}} & \multicolumn{1}{c|}{\textbf{Loss}}& \multicolumn{7}{c}{\textbf{Metrics}} \\
    Pre-trained & RVLoRA & Res-DSC & Pre-trained & RVLoRA & Res-DSC & $L_{mes}$ & Abs Rel$\downarrow$ & Sq Rel$\downarrow$ & RMSE$\downarrow$ & RMSE log$\downarrow$ &$\delta1 \uparrow$&$\delta2 \uparrow$&$\delta3 \uparrow$ \\ \hline
        \ding{53} & \ding{53} & \ding{53}& \ding{53}& \ding{53}& \ding{53}& \ding{53}&   0.058  &   0.430  &   4.808  &   0.080  &   0.972  &   0.998  &   0.999 \\ 
        \checkmark & \ding{53} & \ding{53}& \ding{53}& \ding{53}& \ding{53}& \ding{53}&   0.085  &   0.915  &   7.182  &   0.116  &   0.936  &   0.991  &   0.996  \\ 
        \checkmark & \checkmark & \ding{53}& \ding{53}& \ding{53}& \ding{53}& \ding{53}&   0.052  &   0.352  &   4.363  &   0.072  &   0.979  &   0.998  &   \textbf{1.000} \\ 
        \checkmark & \checkmark & \checkmark & \ding{53}& \ding{53}& \ding{53}& \ding{53}&   0.051  &   0.335  &   4.305  &   0.071  &   \textbf{0.982}  &   0.998  &   \textbf{1.000} \\
        \checkmark & \checkmark & \checkmark & \checkmark& \ding{53}& \ding{53}& \ding{53}&   0.055  &   0.383  &   4.561  &   0.077  &   0.972  &   0.998  &   \textbf{1.000}  \\
        \checkmark & \checkmark & \checkmark & \checkmark& \checkmark& \ding{53}& \ding{53}&   0.051  &   0.329  &   4.199  &   \textbf{0.070}  &   0.981  &   0.998  &   \textbf{1.000} \\
        \checkmark & \checkmark & \checkmark & \checkmark& \checkmark& \checkmark& \ding{53}&   0.055  &   0.414  &   4.742  &   0.077  &   0.975  &   0.997  &   \textbf{1.000} \\
        \checkmark & \checkmark & \checkmark & \checkmark& \checkmark& \ding{53}& \checkmark&   \textbf{0.050}  &   \textbf{0.317}  &   \textbf{4.141}  &   \textbf{0.070} &   \textbf{0.982}  &   \textbf{0.999}  &   \textbf{1.000} \\
        \hline
     \end{tabular}
    } 
   
\end{table*}

\begin{table*}[t]
  \centering
    \caption{The comparison of LoRA methods, vector initialization and layers setting. The best results are presented in bold.}
    \label{tablora}
    \resizebox{0.95\textwidth}{!}{
    \begin{tabular}{c c |c c c c c c c} 
    \hline
    \multicolumn{2}{c|}{} & Abs Rel$\downarrow$ & Sq Rel$\downarrow$ & RMSE$\downarrow$ & RMSE log$\downarrow$ &$\delta1 \uparrow$&$\delta2 \uparrow$&$\delta3 \uparrow$ \\ \hline
        \multirow{4}{*}{LoRA methods} 
        & LoRA \cite{hu2022lora} &   \textbf{0.050}  &   0.321  &   4.182  &   \textbf{0.070}  &   0.980  &   0.998  &   \textbf{1.000}  \\
        & VeRA \cite{kopiczko2024vera}  &   0.056  &   0.414  &   4.726  &   0.078  &   0.972  &   0.997  &   0.999 \\ 
        & DV-LoRA \cite{cui2024endodac} &   0.052  &   0.346  &   4.301  &   0.072  &   0.980  &   0.998  &   \textbf{1.000} \\ 
        & RVLoRA(Ours) &\textbf{0.050} &\textbf{0.317} & \textbf{4.141} & \textbf{0.070} & \textbf{0.982} & \textbf{0.999}  & \textbf{1.000} \\
        \hline
        \multirow{3}{*}{$a$, $b$ Init.} 
        & Uniform &   0.054  &   0.395  &   4.634  &   0.076  &   0.979 &   0.997  &   \textbf{1.000} \\
        & Kaiming Normal &   0.052  &   0.349  &   4.336  &   0.073  &   0.975 &   0.998  &   \textbf{1.000}\\ 
        & Kaiming Uniform(Ours) &   \textbf{0.050}  &   \textbf{0.317}  &   \textbf{4.141}  &   \textbf{0.070}  &   \textbf{0.982} &   \textbf{0.999}  &   \textbf{1.000}\\ 
        \hline
        \multirow{3}{*}{Layers setting}
        & Attention only &   0.052  &   0.353  &   4.426  &   0.073  &   0.975  &   0.998  &  \textbf{1.000} \\
        & Attention \& MLPs &   0.052  &   0.337  &   4.224  &   0.072  &   0.977  &   \textbf{0.999} &  \textbf{1.000}\\ 
        & MLPs  only(Ours) &   \textbf{0.050}  &   \textbf{0.317}  &   \textbf{4.141}  &   \textbf{0.070}  &   \textbf{0.982} &   \textbf{0.999}  &   \textbf{1.000}\\ 
        \hline
     \end{tabular}
    } 
   
\end{table*}

\begin{table*}[t]
  \centering
    \caption{Comparative experimental results for different pre-trained models tested on the SERV-CT and EndoNeRF dataset. The best results are presented in bold.}
    \label{tabpretrainedlora}
    \resizebox{0.85\textwidth}{!}{
    \begin{tabular}{c c |c c c c c c c} 
    \hline
    \multicolumn{2}{c|}{} & Abs Rel$\downarrow$ & Sq Rel$\downarrow$ & RMSE$\downarrow$ & RMSE log$\downarrow$ &$\delta1 \uparrow$&$\delta2 \uparrow$&$\delta3 \uparrow$ \\ \hline
        \multirow{2}{*}{SERV-CT} 
        & LoRA \cite{hu2022lora} & 0.103 & 1.620 & 10.923 & 0.129 & 0.902 & 0.989 &   \textbf{1.000}  \\
        & RVLoRA(Ours) &\textbf{0.072} &\textbf{0.826} & \textbf{8.098} & \textbf{0.095} & \textbf{0.962} & \textbf{0.999}  & \textbf{1.000} \\
        \hline
        \multirow{2}{*}{EndoNeRF} 
        & LoRA \cite{hu2022lora} & 0.176 & 3.467 & 17.014 & 0.216 & 0.678 & 0.973 & \textbf{0.997}  \\
        & RVLoRA(Ours) & \textbf{0.163} & \textbf{2.926} & \textbf{15.818} & \textbf{0.207} & \textbf{0.728} & \textbf{0.976} & \textbf{0.997} \\
        \hline
     \end{tabular}
    } 
   
\end{table*}

While various LoRA fine-tuning methods have been proposed across both natural and medical imaging domains, their efficacy varies. To verify the effectiveness of the designed RVLoRA for the endoscopic monocular depth estimation task, Table \ref{tablora} compares the results of different LoRA fine-tuning methods. It is evident that RVLoRA achieves the best performance across all evaluation metrics. This confirms that the RVLoRA strategy more effectively enhances the model's adaptability for endoscopic depth estimation compared to other LoRA methods. By introducing frozen random vectors to scale the low-rank matrices, the model is compelled to learn essential, scale-robust features rather than overfitting to specific training scales, thereby yielding optimal performance. Specific to the random vectors therein, the initialization method also impacts the results. Table \ref{tablora} showcases the results of the ablation study on different initialization methods for the random vectors $a$ and $b$ in RVLoRA. It demonstrates the effectiveness of Kaiming Uniform initialization compared to Uniform and Kaiming Normal initialization methods for frozen scaling parameters. Further experiments are conducted to verify which specific layer in the transformer block is most effective for applying RVLoRA. The results in Table \ref{tablora} show that placing the RVLoRA at MLP layers can effectively learn image features with a minimal number of trainable parameters and achieve the best depth estimation. 

It was noted that in the direct comparison between LoRA methods, the performance gain of our RVLoRA over standard LoRA was marginal. However, the primary motivation for introducing random vectors in RVLoRA was specifically to enhance the model's generalization and its ability to adapt to scale variations in unseen environments. To validate this, a cross-dataset evaluation was conducted, testing the models trained on the SCARED dataset and directly applying them to the SERV-CT and EndoNeRF datasets. As shown in Table \ref{tabpretrainedlora}, RVLoRA significantly outperforms LoRA on both datasets, confirming that our fine-tuning strategy provides superior scene adaptability and robustness.

\begin{table*}[t]
  \centering
    \caption{The ablation study on the rank setting for RVLoRA on the SCARED dataset with 4 Res-DSC blocks are incorporated after the 3rd, 6th, 9th, and 12th transformer blocks. The best results are presented in bold.}
    \label{tabrankablation}
    \resizebox{0.75\textwidth}{!}{
    \begin{tabular}{c |c c c c c c c} 
    \hline
    Rank for RVLoRA & Abs Rel$\downarrow$ & Sq Rel$\downarrow$ & RMSE$\downarrow$ & RMSE log$\downarrow$ &$\delta1 \uparrow$&$\delta2 \uparrow$&$\delta3 \uparrow$ \\ \hline
        2 & 0.060 & 0.503 & 5.248 & 0.085 & 0.968 & 0.995 &   0.999  \\
        4 &\textbf{0.050} &\textbf{0.317} & \textbf{4.141} & \textbf{0.070} & \textbf{0.982} & \textbf{0.999}  & \textbf{1.000} \\
        8 & 0.052 & 0.327 & 4.168 & 0.071 & 0.978 & \textbf{0.999}  & \textbf{1.000}  \\
        \hline
     \end{tabular}
    } 
   
\end{table*}

\begin{table*}[t]
  \centering
    \caption{The ablation study on the number of Res-DSC on the SCARED dataset with the rank for RVLoRA fixed to 4. The best results are presented in bold.}
    \label{tabresdscablation}
    \resizebox{1.0\textwidth}{!}{
    \begin{tabular}{c |c c c c c c c} 
    \hline
    Num of Res-DSC & Abs Rel$\downarrow$ & Sq Rel$\downarrow$ & RMSE$\downarrow$ & RMSE log$\downarrow$ &$\delta1 \uparrow$&$\delta2 \uparrow$&$\delta3 \uparrow$ \\ \hline
        2 (after the 6th and 12th transformer blocks) & 0.056 & 0.386 & 4.582 & 0.077 & 0.975 & 0.997 &   \textbf{1.000}  \\
        4 (after the 3rd, 6th, 9th, 12th transformer blocks) &\textbf{0.050} &\textbf{0.317} & \textbf{4.141} & \textbf{0.070} & \textbf{0.982} & \textbf{0.999}  & \textbf{1.000} \\
        12 (after all 12 transformer blocks) & 0.053 & 0.366 & 4.394 & 0.075 & 0.970 & 0.998 & \textbf{1.000}  \\
        \hline
     \end{tabular}
    } 
   
\end{table*}

\begin{table*}[t]
  \centering
    \caption{The ablation study on the weight of $L_{mes}$ on the SCARED dataset. The best results are presented in bold.}
    \label{tabmasklossweight}
    \resizebox{0.65\textwidth}{!}{
    \begin{tabular}{c |c c c c c c c} 
    \hline
    $\lambda_{mes} $ & Abs Rel$\downarrow$ & Sq Rel$\downarrow$ & RMSE$\downarrow$ & RMSE log$\downarrow$ &$\delta1 \uparrow$&$\delta2 \uparrow$&$\delta3 \uparrow$ \\ \hline
        0.001 & 0.051 & 0.326 & 4.198 & \textbf{0.070} & 0.977 & \textbf{0.999}  & \textbf{1.000}  \\
        0.003 &\textbf{0.050} &\textbf{0.317} & \textbf{4.141} & \textbf{0.070} & \textbf{0.982} & \textbf{0.999}  & \textbf{1.000} \\
        0.01 & \textbf{0.050} & 0.330 & 4.223 & \textbf{0.070} & 0.981 & 0.998 & \textbf{1.000}  \\
        \hline
     \end{tabular}
    } 
   
\end{table*}

Tables \ref{tabrankablation} and \ref{tabresdscablation} further evaluate the impact of the RVLoRA rank and the number of inserted Res-DSC modules. By varying the rank while fixing the Res-DSC number to 4, it is observed that rank = 4 achieves the best overall performance, whereas lower or higher ranks lead to inferior results. Similarly, when adjusting the number of Res-DSC modules with a fixed rank of 4, uniformly inserting 4 Res-DSC modules into the 12 transformer blocks of the encoder yields the optimal performance. Accordingly, the number of Res-DSC modules and the RVLoRA rank are both set to 4 in the final model.

Table \ref{tabmasklossweight} further evaluates the sensitivity of the mask-guided loss weight $\lambda_{mes} $. The results show that $\lambda = 0.003 $ achieves the best overall performance across most metrics, while smaller or larger weights lead to slight performance degradation. This trend indicates a moderate sensitivity to the loss weight, with performance varying smoothly rather than drastically across different settings. This confirms that the selected weight provides the best performance in our final model.

\begin{table}[!t]
  \centering
    \caption{The results of pose estimation. The best results are presented in bold.}
    \label{tabpose}
    \resizebox{0.5\textwidth}{!}{
    \begin{tabular}{c | c c c} 
    \hline
    Method	& ATE (Seq 1)$\downarrow$ & ATE (Seq 2)$\downarrow$ & ATE (Seq 3)$\downarrow$\\
    \hline 
      AF-SfMLearner \cite{shao2022self} &0.0440 &0.0558 &0.0544\\
      MonoPCC \cite{wang2025monopcc}&0.0596 &0.0617 &0.0825\\
      EndoDAC \cite{cui2024endodac}&0.0439 &0.0577 &0.0529\\
      Ours  &\textbf{0.0424} &\textbf{0.0537} &\textbf{0.0516}\\
    \hline
    \end{tabular}
    }
\end{table}

\begin{figure*}
\centerline{\includegraphics[width=0.8\linewidth]{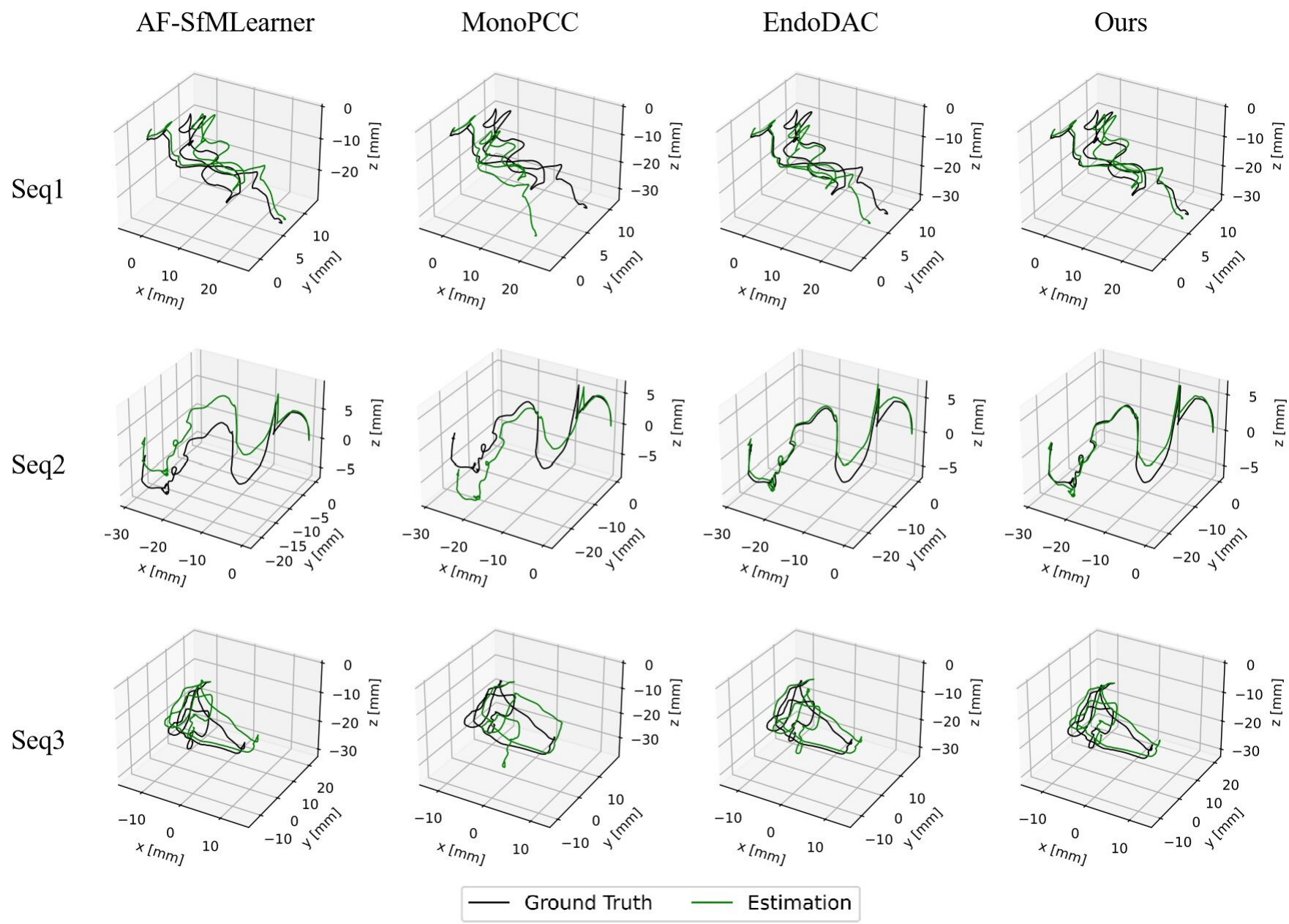}}
\caption{Visualization of camera pose estimation trajectories on the SCARED dataset. The black curve represents the ground truth trajectory. The green curves show the estimated trajectories from different methods.}
\label{fig6}
\end{figure*}

\subsection{Pose estimation}
The self-motion estimation performance of our method was evaluated against other approaches using three image sequences from the SCARED dataset. Table \ref{tabpose} presents the quantitative comparison results with highly competitive methods, AF-SfMLearner, MonoPCC, and EndoDAC. Our method achieves optimal performance across all three sequences, representing an average improvement of 3.21\% over the second-best method.
Fig. \ref{fig6} illustrates the predicted trajectories for three sequences from different methods, with all trajectories aligned at their starting points. Our method's trajectory is consistently the closest to the ground truth (represented by the black curve), exhibiting minimal trajectory drift. This demonstrates the superior performance of the proposed approach in pose estimation.

\section{Discussion}

\begin{figure*}
\centerline{\includegraphics[width=0.45\linewidth]{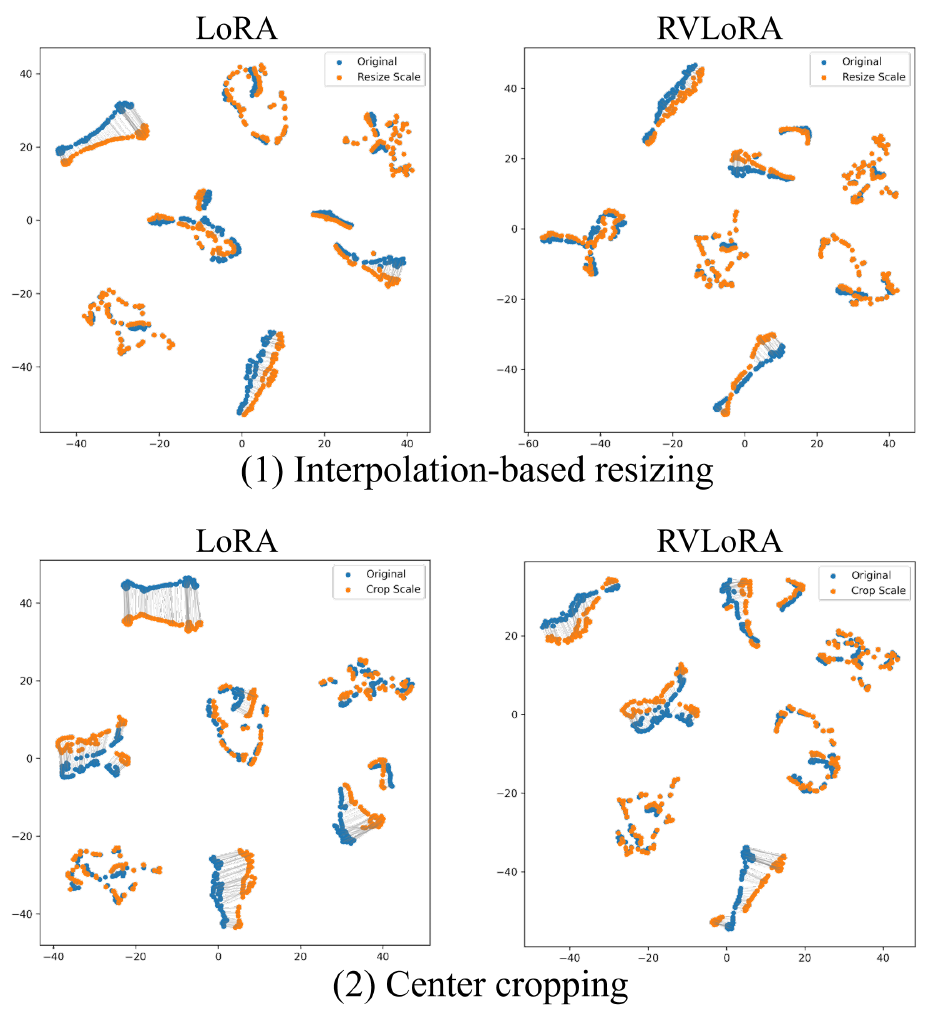}}
\caption{Feature manifold t-SNE visualization under visual scale perturbations. Each gray dashed line connects features extracted from the same image before and after scale change.}
\label{figtsne}
\end{figure*}

\subsection{Effectiveness of RVLoRA}
For further analysis, the effectiveness of RVLoRA can be interpreted from a feature manifold perspective. In monocular depth estimation, scale ambiguity arises when visual scale changes cause excessive displacement in the latent feature space, leading the model to misinterpret scale variations. Multiple variants of images were generated from the SCARED test dataset using two complementary transformations designed to simulate physical scale changes and blur effects: (1) interpolation-based resizing, and (2) center cropping. The t-SNE visualization of the feature manifold, as presented in Fig. \ref{figtsne}, shows that standard LoRA fine-tuning exhibits significant feature drift under scale perturbations, whereas RVLoRA maps multi-scale variants of the same image to compact and coherent regions in the feature space, indicating improved scale-robust representation learning. This behavior fundamentally differs from generic regularization methods such as dropout. The frozen random vectors in RVLoRA impose continuous and structured multiplicative perturbations on low-rank weight updates. These perturbations implicitly simulate the affine transformations in feature responses, encouraging the network to align representations across scales. The improvement is further reflected in zero-shot cross-dataset generalization shown in Table \ref{tabpretrainedlora}, where RVLoRA consistently outperforms standard LoRA under unseen physical scale variations. Together, these observations suggest that RVLoRA alleviates monocular scale ambiguity by enabling structured and scale-robust representation learning.

\begin{figure*}
\centerline{\includegraphics[width=0.55\linewidth]{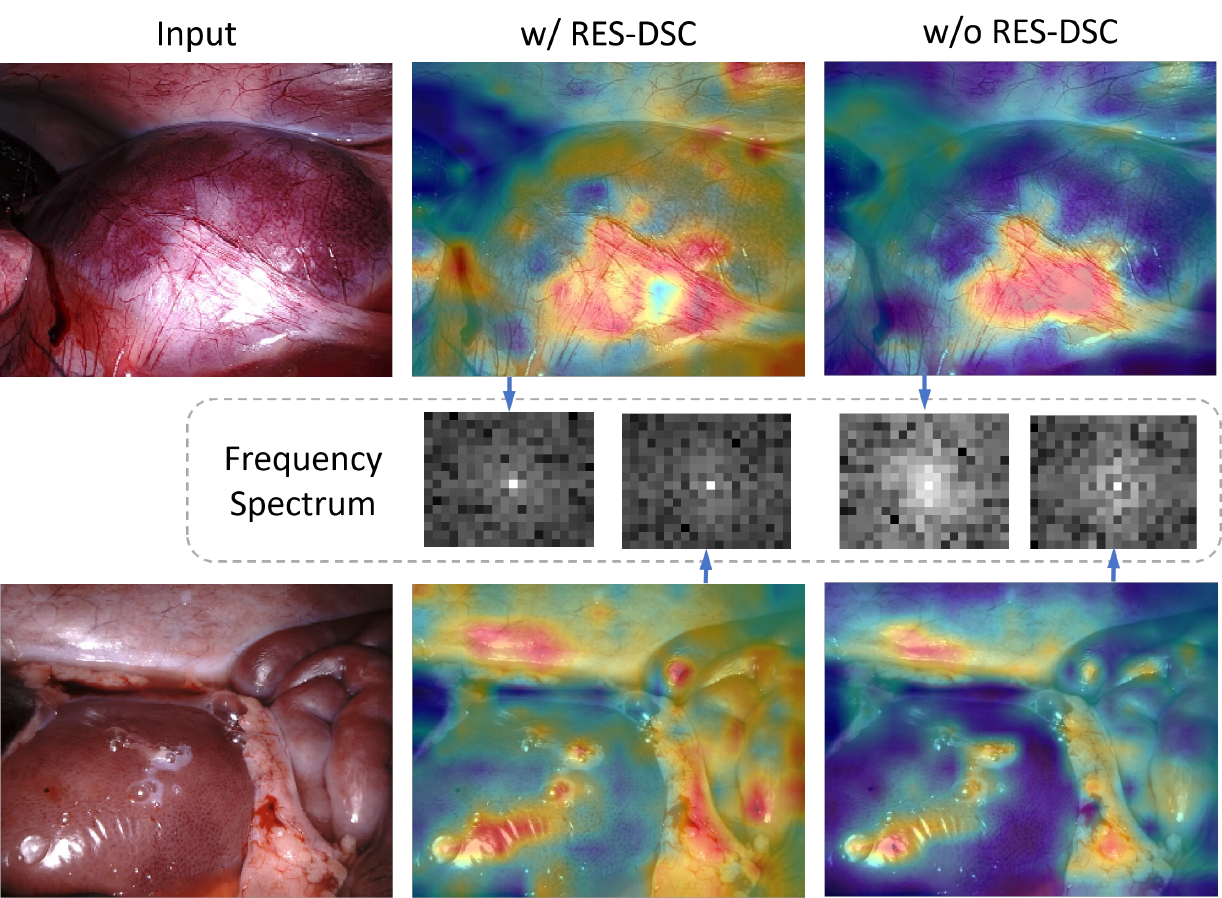}}
\caption{The visualization of the spatial attention and frequency spectrum of feature space.}
\label{figresfrequency}
\end{figure*}

\subsection{Contradictory effects of Res-DSC}

The ablation study shows that introducing the Res-DSC module into the MedSAM branch leads to a degradation in performance. To better understand this phenomenon, a frequency-domain analysis was conducted, as shown in Fig. \ref{figresfrequency}. The results suggest a conflict in inductive biases. The MedSAM (ViT-based) encoder is effective at preserving global semantic structures along with necessary fine-grained texture cues. In contrast, the convolutional inductive bias introduced by Res-DSC acts as a low-pass filter in this branch, which over-smooths the feature representations and suppresses discriminative details. Importantly, this effect is task-dependent. In the depth estimation branch, local feature cues and edge information are beneficial, and thus the convolutional bias of Res-DSC aligns well with task requirements, resulting in consistent performance gains. This analysis provides a principled explanation for the contrasting effects observed in the ablation study and motivates the selective placement of Res-DSC in our final architecture.

\subsection{Robustness of the mask-guided loss design}

Although MedSAM is employed in automatic mode without prompts, its zero-shot segmentation in surgical scenes may contain boundary inaccuracies or over-segmentation due to specular reflections and unstructured tissues. Importantly, the proposed mask-guided edge-aware smoothness loss operates independently within each semantic region and does not explicitly enforce discontinuities across mask borders. Moreover, the edge-aware weighting term suppresses smoothness penalties at strong photometric gradients, allowing true image edges to dominate over potentially noisy mask boundaries. 

\begin{figure*}
\centerline{\includegraphics[width=0.48\linewidth]{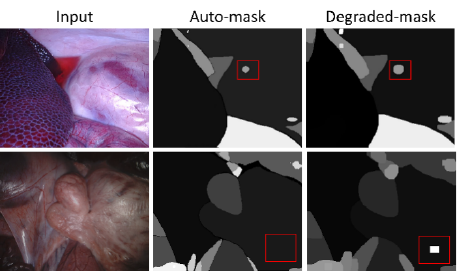}}
\caption{Automatically generated masks and the artificially degraded masks with injected perturbations. The highlighted regions demonstrate the applied morphological perturbations and block-wise noise.}
\label{figdegradedmask}
\end{figure*}

To quantitatively address the risk of error propagation, a sensitivity analysis was conducted by intentionally degrading the mask quality during training. Specific errors like "poor boundary adherence" and "over-segmentation" were simulated by applying random morphological perturbations (dilation/erosion) and adding random block-wise noise (simulating false segments from specular highlights) to the MedSAM masks. Fig. \ref{figdegradedmask} illustrates the original automatically generated masks and the artificially degraded masks with injected perturbations.

The results of retraining the model with degraded masks are presented in Table \ref{tabmaskrobust}. Even when masks are intentionally degraded with morphological perturbations and block-wise noise, incorporating the loss still improves performance compared to removing it entirely. This indicates that the framework remains stable under imperfect segmentation and does not suffer from significant error propagation. The results confirm that the mask guidance serves as a robust auxiliary cue to enhance the overall depth estimation performance.

\begin{table*}[t]
  \centering
    \caption{Comparative experimental results of different mask qualities on the SCARED dataset. The best results are presented in bold.}
    \label{tabmaskrobust}
    \resizebox{0.75\textwidth}{!}{
        \begin{threeparttable}
        \begin{tabular}{c |c c c c c c c} 
        \hline
         & Abs Rel$\downarrow$ & Sq Rel$\downarrow$ & RMSE$\downarrow$ & RMSE log$\downarrow$ &$\delta1 \uparrow$&$\delta2 \uparrow$&$\delta3 \uparrow$ \\ \hline
            w/o $\lambda_{mes} $ & 0.051 & 0.329 & 4.199 & \textbf{0.070} & 0.981 & 0.998  & \textbf{1.000}  \\
            w/ $\lambda_{mes} $ \tnote{*} &0.051 &0.326 & 4.195 & \textbf{0.070} & \textbf{0.982} & \textbf{0.999}  & \textbf{1.000} \\
             w/ $\lambda_{mes} $ (Ours) &\textbf{0.050} &\textbf{0.317} & \textbf{4.141} & \textbf{0.070} & \textbf{0.982} & \textbf{0.999}  & \textbf{1.000}  \\
            \hline
         \end{tabular}
         \begin{tablenotes}
          \footnotesize
          \item[*] Random morphological perturbations and block-wise noise are introduced to the auto-generated masks.
        \end{tablenotes}
        \end{threeparttable}
    } 
\end{table*}

\section{Conclusion}
In this work, EndoUFM is introduced as a novel self-supervised framework that utilizes foundation models to enhance depth estimation accuracy and robustness in challenging endoscopic scenes. Our method leverages large-scale, pre-learned knowledge by integrating priors from the Depth Anything Model and the Segment Anything Model, specifically MedSAM, providing a robust foundation for understanding complex surgical structures. An adaptive fine-tuning strategy, combining RV-LoRA and Res-DSC, is designed to emphasize model adaptability and the capture of fine-grained features. Furthermore, the mask-guided smoothness loss, based on MedSAM's pre-segmented masks, effectively enhances semantic consistency in the depth predictions. Extensive experiments on four benchmark datasets demonstrate that our method achieves state-of-the-art performance and exhibits exceptional robustness in challenging scenarios. Despite the excellent performance, our method still exhibits some limitations in areas with uneven illumination or low brightness. Therefore, 3D scene perception and uncertainty estimation will be incorporated in future work.

In conclusion, the EndoUFM framework establishes a new state-of-the-art in unsupervised monocular depth estimation for endoscopic scenes. It provides reliable point cloud initialization for subsequent 3D reconstruction using methods like NeRF \cite{mildenhall2021nerf} and 3D Gaussian Splatting \cite{kerbl20233d}. Ultimately, these advancements represent a crucial step toward creating more sophisticated Augmented Reality (AR) and surgical navigation systems, with the potential to significantly enhance procedural precision and safety in minimally invasive surgery.






{
\bibliographystyle{elsarticle-num}

\bibliography{root_f}

@article{peng2025hadepth,
Author = {Peng, Xiaoyuan and Wang, Shigang and Wei, Jian and Zhao, Yan},
Title = {HADepth: Highlight-aware monocular depth estimation for endoscopy},
Journal = {SIGNAL IMAGE AND VIDEO PROCESSING},
Year = {2025},
Volume = {19},
Number = {10},
Month = {JUN 26},
Article-Number = {810},
ISSN = {1863-1703},
EISSN = {1863-1711},
}

@inproceedings{Bian2019unsupervisedscale,
 author = {Bian, Jiawang and Li, Zhichao and Wang, Naiyan and Zhan, Huangying and Shen, Chunhua and Cheng, Ming-Ming and Reid, Ian},
 booktitle = {Advances in Neural Information Processing Systems},
 editor = {H. Wallach and H. Larochelle and A. Beygelzimer and F. d\textquotesingle Alch\'{e}-Buc and E. Fox and R. Garnett},
 pages = {},
 publisher = {Curran Associates, Inc.},
 title = {Unsupervised Scale-consistent Depth and Ego-motion Learning from Monocular Video},
 volume = {32},
 year = {2019}
}

@article{Song2024unsupervisedconsistency,
  author={Song, Xiaogang and Hu, Haoyue and Liang, Li and Shi, Weiwei and Xie, Guo and Lu, Xiaofeng and Hei, Xinhong},
  journal={IEEE Transactions on Multimedia}, 
  title={Unsupervised Monocular Estimation of Depth and Visual Odometry Using Attention and Depth-Pose Consistency Loss}, 
  year={2024},
  volume={26},
  number={},
  pages={3517-3529},
}

@article{ZHANG2026Multi,
title = {Multi-Source Temporal-Depth fusion for robust end-to-End visual odometry},
journal = {Neural Networks},
volume = {198},
pages = {108598},
year = {2026},
issn = {0893-6080},
author = {Sihang Zhang and Congqi Cao and Qiang Gao and Ganchao Liu}
}

@article{sharma2025utilization,
  title={The Utilization of Navigation and Emerging Technologies With Endoscopic Spine Surgery: A Narrative Review},
  author={Sharma, Abhinav K and de Oliveira, Rafael Garcia and Suvithayasiri, Siravich and Chavalparit, Piya and Chang, Chien Chun and Kim, Yong H and Fischer, Charla R and Lee, Sang and Cho, Samuel and Kim, Jin-Sung and others},
  journal={Neurospine},
  volume={22},
  number={1},
  pages={105},
  year={2025}
}

@article{yang2024selflightweight,
  title={Self-supervised lightweight depth estimation in endoscopy combining cnn and transformer},
  author={Yang, Zhuoyue and Pan, Junjun and Dai, Ju and Sun, Zhen and Xiao, Yi},
  journal={IEEE Transactions on Medical Imaging},
  volume={43},
  number={5},
  pages={1934--1944},
  year={2024},
  publisher={IEEE}
}

@inproceedings{liu2024self,
  title={Self-supervised monocular depth estimation with effective feature fusion and self distillation},
  author={Liu, Zhenfei and Song, Chengqun and Cheng, Jun and Luo, Jiefu and Wang, Xiaoyang},
  booktitle={2024 IEEE/RSJ International Conference on Intelligent Robots and Systems (IROS)},
  pages={7160--7166},
  year={2024},
  organization={IEEE}
}

@inproceedings{zhou2017unsupervised,
  title={Unsupervised learning of depth and ego-motion from video},
  author={Zhou, Tinghui and Brown, Matthew and Snavely, Noah and Lowe, David G},
  booktitle={Proceedings of the IEEE conference on computer vision and pattern recognition},
  pages={1851--1858},
  year={2017}
}

@article{ozyoruk2021endoslam,
  title={EndoSLAM dataset and an unsupervised monocular visual odometry and depth estimation approach for endoscopic videos},
  author={Ozyoruk, Kutsev Bengisu and Gokceler, Guliz Irem and Bobrow, Taylor L and Coskun, Gulfize and Incetan, Kagan and Almalioglu, Yasin and Mahmood, Faisal and Curto, Eva and Perdigoto, Luis and Oliveira, Marina and others},
  journal={Medical image analysis},
  volume={71},
  pages={102058},
  year={2021},
  publisher={Elsevier}
}

@article{li2024image,
  title={Image intrinsic-based unsupervised monocular depth estimation in endoscopy},
  author={Li, Bojian and Liu, Bo and Zhu, Miao and Luo, Xiaoyan and Zhou, Fugen},
  journal={IEEE Journal of Biomedical and Health Informatics},
  year={2024},
  publisher={IEEE}
}

@article{shao2022self,
  title={Self-supervised monocular depth and ego-motion estimation in endoscopy: Appearance flow to the rescue},
  author={Shao, Shuwei and Pei, Zhongcai and Chen, Weihai and Zhu, Wentao and Wu, Xingming and Sun, Dianmin and Zhang, Baochang},
  journal={Medical image analysis},
  volume={77},
  pages={102338},
  year={2022},
  publisher={Elsevier}
}

@inproceedings{godard2019digging,
  title={Digging into self-supervised monocular depth estimation},
  author={Godard, Cl{\'e}ment and Mac Aodha, Oisin and Firman, Michael and Brostow, Gabriel J},
  booktitle={Proceedings of the IEEE/CVF international conference on computer vision},
  pages={3828--3838},
  year={2019}
}

@article{han2024depth,
  title={Depth anything in medical images: A comparative study},
  author={Han, John J and Acar, Ayberk and Henry, Callahan and Wu, Jie Ying},
  journal={arXiv preprint arXiv:2401.16600},
  year={2024}
}

@inproceedings{yang2024depth,
  title={Depth anything: Unleashing the power of large-scale unlabeled data},
  author={Yang, Lihe and Kang, Bingyi and Huang, Zilong and Xu, Xiaogang and Feng, Jiashi and Zhao, Hengshuang},
  booktitle={Proceedings of the IEEE/CVF conference on computer vision and pattern recognition},
  pages={10371--10381},
  year={2024}
}

@article{ma2024segment,
  title={Segment anything in medical images},
  author={Ma, Jun and He, Yuting and Li, Feifei and Han, Lin and You, Chenyu and Wang, Bo},
  journal={Nature Communications},
  volume={15},
  number={1},
  pages={654},
  year={2024},
  publisher={Nature Publishing Group UK London}
}

@inproceedings{
hu2022lora,
title={Lo{RA}: Low-Rank Adaptation of Large Language Models},
author={Edward J Hu and yelong shen and Phillip Wallis and Zeyuan Allen-Zhu and Yuanzhi Li and Shean Wang and Lu Wang and Weizhu Chen},
booktitle={International Conference on Learning Representations},
year={2022}
}

@article{wang2025monopcc,
  title={MonoPCC: Photometric-invariant cycle constraint for monocular depth estimation of endoscopic images},
  author={Wang, Zhiwei and Zhou, Ying and He, Shiquan and Li, Ting and Huang, Fan and Ding, Qiang and Feng, Xinxia and Liu, Mei and Li, Qiang},
  journal={Medical Image Analysis},
  volume={102},
  pages={103534},
  year={2025},
  publisher={Elsevier}
}

@inproceedings{batlle2022photometric,
  title={Photometric single-view dense 3D reconstruction in endoscopy},
  author={Batlle, Victor M and Montiel, Jos{\'e} MM and Tard{\'o}s, Juan D},
  booktitle={2022 IEEE/RSJ International Conference on Intelligent Robots and Systems (IROS)},
  pages={4904--4910},
  year={2022},
  organization={IEEE}
}

@article{ranftl2020towards,
  title={Towards robust monocular depth estimation: Mixing datasets for zero-shot cross-dataset transfer},
  author={Ranftl, Ren{\'e} and Lasinger, Katrin and Hafner, David and Schindler, Konrad and Koltun, Vladlen},
  journal={IEEE transactions on pattern analysis and machine intelligence},
  volume={44},
  number={3},
  pages={1623--1637},
  year={2020},
  publisher={IEEE}
}

@article{oquab2024dinov2,
  title={DINOv2: Learning Robust Visual Features without Supervision},
  author={Oquab, Maxime and Darcet, Timoth{\'e}e and Moutakanni, Th{\'e}o and Vo, Huy and Szafraniec, Marc and Khalidov, Vasil and Fernandez, Pierre and Haziza, Daniel and Massa, Francisco and El-Nouby, Alaaeldin and others},
  journal={Transactions on Machine Learning Research Journal},
  pages={1--31},
  year={2024}
}

@article{tian2024endoomni,
  title={EndoOmni: Zero-shot cross-dataset depth estimation in endoscopy by robust self-learning from noisy labels},
  author={Tian, Qingyao and Chen, Zhen and Liao, Huai and Huang, Xinyan and Li, Lujie and Ourselin, Sebastien and Liu, Hongbin},
  journal={arXiv preprint arXiv:2409.05442},
  year={2024}
}

@article{cui2024surgical,
  title={Surgical-dino: adapter learning of foundation models for depth estimation in endoscopic surgery},
  author={Cui, Beilei and Islam, Mobarakol and Bai, Long and Ren, Hongliang},
  journal={International Journal of Computer Assisted Radiology and Surgery},
  volume={19},
  number={6},
  pages={1013--1020},
  year={2024},
  publisher={Springer}
}

@inproceedings{cui2024endodac,
  title={Endodac: Efficient adapting foundation model for self-supervised depth estimation from any endoscopic camera},
  author={\vspace{0mm}Cui, Beilei and Islam, Mobarakol and Bai, Long and Wang, An and Ren, Hongliang},
  booktitle={International Conference on Medical Image Computing and Computer-Assisted Intervention},
  pages={208--218},
  year={2024},
  organization={Springer}
}

@article{yang2024selfCLIP,
  title={Self-supervised endoscopy depth estimation framework with CLIP-guidance segmentation},
  author={Yang, Zhuoyue and Pan, Junjun and Dai, Ju and Sun, Zhen and Xiao, Yi},
  journal={Biomedical Signal Processing and Control},
  volume={95},
  pages={106410},
  year={2024},
  publisher={Elsevier}
}

@inproceedings{radford2021learning,
  title={Learning transferable visual models from natural language supervision},
  author={Radford, Alec and Kim, Jong Wook and Hallacy, Chris and Ramesh, Aditya and Goh, Gabriel and Agarwal, Sandhini and Sastry, Girish and Askell, Amanda and Mishkin, Pamela and Clark, Jack and others},
  booktitle={International conference on machine learning},
  pages={8748--8763},
  year={2021},
  organization={PmLR}
}

@inproceedings{hayou2024lora+,
  title={LoRA+: Efficient Low Rank Adaptation of Large Models},
  author={Hayou, Soufiane and Ghosh, Nikhil and Yu, Bin},
  booktitle={International Conference on Machine Learning},
  pages={17783--17806},
  year={2024},
  organization={PMLR}
}

@inproceedings{zhang2023adaptive,
  title={ADAPTIVE BUDGET ALLOCATION FOR PARAMETER-EFFICIENT FINE-TUNING},
  author={Zhang, Qingru and Chen, Minshuo and Bukharin, Alexander and He, Pengcheng and Cheng, Yu and Chen, Weizhu and Zhao, Tuo},
  booktitle={11th International Conference on Learning Representations, ICLR 2023},
  year={2023}
}

@inproceedings{
kopiczko2024vera,
title={Ve{RA}: Vector-based Random Matrix Adaptation},
author={Dawid Jan Kopiczko and Tijmen Blankevoort and Yuki M Asano},
booktitle={12th International Conference on Learning Representations},
year={2024}
}

@inproceedings{kirillov2023segment,
  title={Segment anything},
  author={Kirillov, Alexander and Mintun, Eric and Ravi, Nikhila and Mao, Hanzi and Rolland, Chloe and Gustafson, Laura and Xiao, Tete and Whitehead, Spencer and Berg, Alexander C and Lo, Wan-Yen and others},
  booktitle={Proceedings of the IEEE/CVF international conference on computer vision},
  pages={4015--4026},
  year={2023}
}

@inproceedings{aghajanyan2021intrinsic,
  title={Intrinsic Dimensionality Explains the Effectiveness of Language Model Fine-Tuning},
  author={Aghajanyan, Armen and Gupta, Sonal and Zettlemoyer, Luke},
  booktitle={Proceedings of the 59th Annual Meeting of the Association for Computational Linguistics and the 11th International Joint Conference on Natural Language Processing (Volume 1: Long Papers)},
  pages={7319--7328},
  year={2021}
}

@inproceedings{lu2022frozen,
  title={Frozen pretrained transformers as universal computation engines},
  author={Lu, Kevin and Grover, Aditya and Abbeel, Pieter and Mordatch, Igor},
  booktitle={Proceedings of the AAAI conference on artificial intelligence},
  volume={36},
  number={7},
  pages={7628--7636},
  year={2022}
}

@inproceedings{
frankle2021training,
title={Training BatchNorm and Only BatchNorm: On the Expressive Power of Random Features in {\{}CNN{\}}s},
author={Jonathan Frankle and David J. Schwab and Ari S. Morcos},
booktitle={International Conference on Learning Representations},
year={2021}
}

@inproceedings{he2015delving,
  title={Delving deep into rectifiers: Surpassing human-level performance on imagenet classification},
  author={He, Kaiming and Zhang, Xiangyu and Ren, Shaoqing and Sun, Jian},
  booktitle={Proceedings of the IEEE international conference on computer vision},
  pages={1026--1034},
  year={2015}
}

@inproceedings{shvets2024joint,
  title={Joint depth prediction and semantic segmentation with multi-view sam},
  author={Shvets, Mykhailo and Zhao, Dongxu and Niethammer, Marc and Sengupta, Roni and Berg, Alexander C},
  booktitle={Proceedings of the IEEE/CVF Winter Conference on Applications of Computer Vision},
  pages={1328--1338},
  year={2024}
}

@inproceedings{woo2018cbam,
  title={Cbam: Convolutional block attention module},
  author={Woo, Sanghyun and Park, Jongchan and Lee, Joon-Young and Kweon, In So},
  booktitle={Proceedings of the European conference on computer vision (ECCV)},
  pages={3--19},
  year={2018}
}

@article{krebs2020intrinsic,
  title={Intrinsic image decomposition as two independent deconvolution problems},
  author={Krebs, Alexandre and Benezeth, Yannick and Marzani, Franck},
  journal={Signal Processing: Image Communication},
  volume={86},
  pages={115872},
  year={2020},
  publisher={Elsevier}
}

@article{allan2021stereo,
  title={Stereo correspondence and reconstruction of endoscopic data challenge},
  author={Allan, Max and Mcleod, Jonathan and Wang, Congcong and Rosenthal, Jean Claude and Hu, Zhenglei and Gard, Niklas and Eisert, Peter and Fu, Ke Xue and Zeffiro, Trevor and Xia, Wenyao and others},
  journal={arXiv preprint arXiv:2101.01133},
  year={2021}
}

@article{mountney2010three,
  title={Three-dimensional tissue deformation recovery and tracking},
  author={Mountney, Peter and Stoyanov, Danail and Yang, Guang-Zhong},
  journal={IEEE Signal Processing Magazine},
  volume={27},
  number={4},
  pages={14--24},
  year={2010},
  publisher={IEEE}
}

@article{edwards2022serv,
  title={SERV-CT: A disparity dataset from cone-beam CT for validation of endoscopic 3D reconstruction},
  author={Edwards, PJ Eddie and Psychogyios, Dimitris and Speidel, Stefanie and Maier-Hein, Lena and Stoyanov, Danail},
  journal={Medical image analysis},
  volume={76},
  pages={102302},
  year={2022},
  publisher={Elsevier}
}

@inproceedings{wang2022neural,
  title={Neural rendering for stereo 3d reconstruction of deformable tissues in robotic surgery},
  author={Wang, Yuehao and Long, Yonghao and Fan, Siu Hin and Dou, Qi},
  booktitle={International conference on medical image computing and computer-assisted intervention},
  pages={431--441},
  year={2022},
  organization={Springer}
}

@article{recasens2021endo,
  title={Endo-depth-and-motion: Reconstruction and tracking in endoscopic videos using depth networks and photometric constraints},
  author={Recasens, David and Lamarca, Jos{\'e} and F{\'a}cil, Jos{\'e} M and Montiel, JM Martinez and Civera, Javier},
  journal={IEEE Robotics and Automation Letters},
  volume={6},
  number={4},
  pages={7225--7232},
  year={2021},
  publisher={IEEE}
}

@inproceedings{li2021revisiting,
  title={Revisiting stereo depth estimation from a sequence-to-sequence perspective with transformers},
  author={Li, Zhaoshuo and Liu, Xingtong and Drenkow, Nathan and Ding, Andy and Creighton, Francis X and Taylor, Russell H and Unberath, Mathias},
  booktitle={Proceedings of the IEEE/CVF international conference on computer vision},
  pages={6197--6206},
  year={2021}
}

@article{paszke2019pytorch,
  title={Pytorch: An imperative style, high-performance deep learning library},
  author={Paszke, Adam and Gross, Sam and Massa, Francisco and Lerer, Adam and Bradbury, James and Chanan, Gregory and Killeen, Trevor and Lin, Zeming and Gimelshein, Natalia and Antiga, Luca and others},
  journal={Advances in neural information processing systems},
  volume={32},
  year={2019}
}

@article{mildenhall2021nerf,
  title={Nerf: Representing scenes as neural radiance fields for view synthesis},
  author={Mildenhall, Ben and Srinivasan, Pratul P and Tancik, Matthew and Barron, Jonathan T and Ramamoorthi, Ravi and Ng, Ren},
  journal={Communications of the ACM},
  volume={65},
  number={1},
  pages={99--106},
  year={2021},
  publisher={ACM New York, NY, USA}
}

@article{kerbl20233d,
  title={3D Gaussian Splatting for Real-Time Radiance Field Rendering},
  author={Kerbl, Bernhard and Kopanas, Georgios and Leimk{\"u}hler, Thomas and Drettakis, George},
  journal={ACM Transactions on Graphics},
  volume={42},
  number={4},
  pages={1--14},
  year={2023}
}

@inproceedings{Kingma2014AdamAM,
  title={Adam: A Method for Stochastic Optimization},
  author={Diederik P. Kingma and Jimmy Ba},
  booktitle={3rd International Conference on Learning Representa
tions, ICLR 2015, ConferenceTrackProceedings},
  year={2015}
}
}

\end{document}